\documentclass[10pt,twocolumn,letterpaper]{article}
\usepackage{iccv}
\usepackage{times}
\usepackage{epsfig}
\usepackage{graphicx}
\usepackage{amsmath}
\usepackage{amssymb}
\usepackage{multirow}
\usepackage{tabu}
\usepackage{color}
\usepackage[T1]{fontenc}
\usepackage[utf8]{inputenc}
\usepackage{authblk}
\usepackage[breaklinks=true,bookmarks=false]{hyperref}
\newcommand{\zhiyin}[1]{{\color{blue}#1}}
\newcommand{\code}[1]{{\color{magenta}#1}}
\newcommand{\myparagraph}[1]{{\setlength{\parskip}{0.3em} \noindent \textbf {#1}}}

\iccvfinalcopy % *** Uncomment this line for the final submission

\def\iccvPaperID{****} % *** Enter the ICCV Paper ID here

\ificcvfinal\fi

\begin{document}

%%%%%%%%% TITLE
% \title{\vspace{-1.5cm}Unified Pre-training with Pseudo Texts for Text-To-Image \\ Person Re-identification}
\title{Unified Pre-training with Pseudo Texts for Text-To-Image \\ Person Re-identification}

\author{Zhiyin Shao\textsuperscript{\rm 1,2\thanks{Equal contribution. 
$^\dag$Corresponding author.
}}, 
Xinyu Zhang\textsuperscript{\rm 2$*$}, 
Changxing Ding\textsuperscript{\rm 1,3$^\dag$}, 
Jian Wang\textsuperscript{\rm 2}, 
Jingdong Wang\textsuperscript{\rm 2}\\
\normalsize{
\textsuperscript{\rm 1}South China University of Technology, China  \textsuperscript{\rm 2}Baidu VIS, China 
\textsuperscript{\rm 3}Pazhou Lab, China
}

{\tt\small eezyshao@mail.scut.edu.cn
}
{\tt\small chxding@scut.edu.cn
}
{\tt\small \{zhangxinyu14,wangjian33,wangjingdong\}@baidu.com
}

}

% \author{Zhiyin Shao\\
% South China University of Technology\\
% Guangzhou\\
% {\tt\small shaozhiyin@baidu.com}
% \and
% Xinyu Zhang\\
% Baidu VIS\\
% Beijing\\
% {\tt\small zhangxinyu14@baidu.com}
% \and
% Changxing Ding\\
% South China University of Technology\\
% Guangzhou\\
% {\tt\small chxding@scut.edu.cn}
% \and
% Jian Wang\\
% Baidu VIS\\
% Beijing\\
% {\tt\small wangjian33@baidu.com}
% \and
% Jingdong Wang\\
% Baidu VIS\\
% Beijing\\
% {\tt\small wangjingdong@outlook.com}
% }

\maketitle
\ificcvfinal\thispagestyle{empty}\fi

%%%%%%%%% ABSTRACT
\begin{abstract}
% The pre-training task is an indispensable precursor to the text-to-image person re-identification (T2I-ReID) task.
The pre-training task is indispensable for the text-to-image person re-identification (T2I-ReID) task.
However, there are two underlying inconsistencies between these two tasks that may impact the performance:
% This paper aims to address the underlying gaps between these two tasks.
% The gaps mainly lie in two aspects:
i) \textbf{Data} inconsistency. 
A large domain gap exists between the generic images/texts used in public pre-trained models and the specific person data in the T2I-ReID task.
This gap is especially severe for texts, as general textual data are usually unable to describe specific people in fine-grained detail.
ii) \textbf{Training} inconsistency. The processes of pre-training of images and texts are independent, despite 
% In contrast, 
cross-modality learning being critical to T2I-ReID.
% These two gaps result in the inconsistency 
To address the above issues, we present a new unified pre-training pipeline (UniPT) designed specifically for the T2I-ReID task.
% text-to-image person Re-ID, named \textbf{VLP-ReID}.
We first build a large-scale text-labeled person dataset ``LUPerson-T'',
%\footnote{Images in ``LUPerson-\textbf{T}'' are from ``LUPerson''~\cite{fu2021unsupervised}, which is a public unlabeled person dataset.}, 
in which
% The corresponding human characters are automatically generated by the CLIP paradigm
pseudo-textual descriptions of images are automatically generated by the CLIP paradigm using a divide-conquer-combine strategy.
% Benefiting from this dataset, we then utilize a simple vision-and-language pre-training framework to explicitly align the feature space of image and text modalities during pre-training.
Benefiting from this dataset, we then utilize a simple vision-and-language pre-training framework to explicitly align the feature space of the image and text modalities during pre-training.
In this way, the pre-training task and the T2I-ReID task are made consistent with each other on both data and training levels.
Without the need for any bells and whistles, our UniPT achieves competitive Rank-1 accuracy of, \ie, 68.50\%, 60.09\%, and 51.85\% on CUHK-PEDES, ICFG-PEDES and RSTPReid, respectively.
Both the LUPerson-T dataset and code are available at \href{}{\code{https://github.com/ZhiyinShao-H/UniPT}}.

\end{abstract}

%%%%%%%%% BODY TEXT
\section{Introduction}
Text-to-image person re-identification~\cite{li2017person} (T2I-ReID) is a retrieval task that aims to search specific person images based on natural language descriptions~\cite{li2017person}. 
Compared with large-scale datasets in general image-text retrieval tasks~\cite{sarafianos2019adversarial,liu2020graph,zhang2022negative}, existing T2I-ReID datasets suffer from limited scale and diversity.
Therefore, the prior knowledge learned through large-scale pre-training is critical if good performance on the T2I-ReID task is to be achieved.

\begin{figure}[t!]
\begin{center}
\includegraphics[width=1\linewidth]{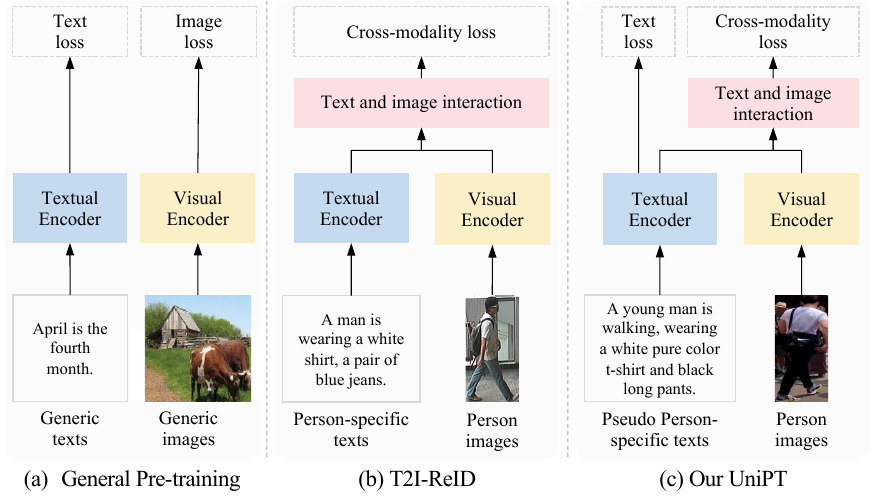}
\end{center}
  \caption{Pipeline comparison on (a) general pre-training in previous works, (b) T2I-ReID tasks, and (c) our unified pre-training (UniPT).
  Our UniPT shares a similar format with the T2I-ReID pipeline at both the data and training levels.
  % between previous methods and our unified architecture. Previous methods employ (a) overall two-stream model or (b) partial two-stream model for visual and textual inputs separately. (c) Our unified architecture build one shared Transformer for both modalities.
  }
  \vspace{-0.2cm}
\label{fig:fig1}
\end{figure}

Previous works~\cite{zheng2020dual, zhang2018deep, wang2019language, liu2019deep,niu2020textual} simply utilize
% fully takes advantages of 
publicly released pre-trained models that have learned from generic data.
In more detail, the visual encoders~\cite{he2016deep, dosovitskiy2020image} are usually pre-trained on ImageNet~\cite{russakovsky2015imagenet}, while the textual encoders~\cite{hochreiter1997long, devlin2018bert} learn on hundreds of millions of words~\cite{zhu2015aligning, devlin2018bert}.
These models naturally serve as the initial parameters of visual and textual encoders in T2I-ReID.
The existing literature~\cite{wu2021lapscore,shao2022learning,farooq2022axm,zheng2020hierarchical,wang2021text,ding2021semantically} thus pays more attention to designing various cross-modality modules to mitigate the modality gap~\cite{wu2021lapscore,shao2022learning,farooq2022axm} and the part-alignment problems~\cite{zheng2020hierarchical,wang2021text,ding2021semantically} instead.
No extant research has yet investigated how the pre-training task affects the following T2I-ReID task.
% is still under explored.

In this paper, we make the first attempt to explicitly reveal the inconsistencies between the existing pre-training task and the T2I-ReID task.
As shown in Figure~\ref{fig:fig1}, there are two main inconsistencies: \ie, data and training.
Regarding the \textit{data inconsistency}, we observe that the pre-training data are obtained from various objects and scenes, while the T2I-ReID data are specific to pedestrians:
a large domain gap exists between these two kinds of datasets.
A similar observation can be made with respect to image-based person ReID~\cite{fu2021unsupervised,luo2021self,fu2022large}.
Accordingly, \cite{fu2021unsupervised} proposes a large-scale person dataset, \ie, ``LUPerson''.
Despite the possibility of initializing the visual encoder with LUPerson pre-trained models, this approach is not optimal due to the remaining textual gap.
In fact, the generic text data fail to describe the person in the kind of fine-grained detail that is crucial to capturing an individual person’s characteristics.
This highlights the requirement of \textit{a large-scale image-text person dataset for the pre-training}. 
% that \textit{how to alleviate the data gap with the similar domain data as T2I-ReID person datasets?}
Moreover, regarding \textit{training inconsistency}, it is worth noting that the initial visual and textual models in previous methods are pre-trained independently.
This approach might not be suitable for T2I-ReID, since the interaction between images and texts key to narrowing down the modality gap.
We therefore consider \textit{how to directly apply the modality interaction during the pre-training to match the process in T2I-ReID.}

To this end, we present a novel unified 
% vision-and-language 
pre-training pipeline, \textbf{UniPT}, which is specifically designed for the T2I-ReID task.
Enabled by LUPerson~\cite{fu2021unsupervised}, we create a new text-labeled variant, \ie, LUPerson-\textbf{T}, for the model pre-training.
Removing the labor and cost associated with manual annotation,
the pseudo-text descriptions are derived automatically using a novel divide-conquer-combine strategy based on CLIP~\cite{radford2021learning}. 
Specifically, we first divide the person characteristics into attribute phrases, \eg, ``a blue striped shirt'' and ``long hair'' (\textit{divide} stage).
Next, we convert these phrases to prompts, and input them into the frozen CLIP model to obtain the most matched attributes for specific person images (\textit{conquer} stage).
Subsequently, we insert the matched attributes into pre-defined templates to obtain complete pseudo-texts. 
Moreover, we adopt synonym substitution to enrich the diversity of descriptions (\textit{combine} stage). 
Despite the noise contained in the pseudo-texts, LUPerson-T focuses describing individual persons in detail, effectively reducing the data domain gap between the pre-training and the T2I-ReID tasks.

Based on the LUPerson-T dataset, we simply use the vision-and-language pre-training framework in the pre-training process.
% as the similar as the T2I-ReID pre-training pipeline.
Inspired by CLIP~\cite{radford2021learning}, we employ the contrastive loss on pairs of person images and their pseudo-texts.
Meanwhile, we apply the masked language model objective~\cite{devlin2018bert} 
% to predict the masked words 
to prevent the underlying over-fitting.
In this way, the features of images and texts can be made to align with each other during the pre-training, ensuring that the training processes of the pre-training and T2I-ReID tasks are consistent.
In summary, our key contributions are as follows:
\begin{itemize}
\itemsep -.151cm
\item We first reveal the data and training inconsistencies between the pre-training and T2I-ReID tasks.
\item We build a text-labeled dataset LUPerson-T, in which the pseudo text descriptions are automatically generated by the proposed divide-conquer-combine strategy.
% \item We propose to use the vision-and-language pre-training pipeline for the T2I-ReID pre-training.
\item We propose a unified pre-training pipeline, namely UniPT, on LUPerson-T to make the pre-training and the T2I-ReID task consistent at both the data and training levels.
% data and training consistent between the pre-training and the T2I-ReID task.
\item We conduct comprehensive experiments and analyses to
show the effectiveness of our UniPT. Our method outperforms current state-of-the-art methods on three benchmarks.
% without bells and whistles.
\end{itemize}
%------------------------------------------------------------------------

% We provide the statistic of the pre-training dataset and the T2I-ReID dataset in Table~\ref{tab:statistic}.

\begin{table}
\centering
\footnotesize
\setlength{\tabcolsep}{0.8mm}{
\begin{tabu} to 0.98\linewidth {l|l|c|c|c|c}
% \begin{tabular}{l|cccccc}
    \hline
     \multicolumn{1}{l|}{\multirow{2}{*}{Task}} & \multicolumn{1}{l|}{\multirow{2}{*}{Dataset}} & \multicolumn{1}{c|}{Person} & \multicolumn{1}{c|}{Image} & \multicolumn{2}{c}{Text} \\
    \cline{4-6}
      &  & \multicolumn{1}{c|}{Only} & Size & Size & Source \\
    \hline\hline
     & \multicolumn{1}{l|}{\multirow{1}{*}{ImageNet~\cite{russakovsky2015imagenet}}} & $\times$ & 1.3M & - & - \\
   \multirow{1}{*}{Pre-} & \multicolumn{1}{l|}{\multirow{1}{*}{Text data in \cite{devlin2018bert}}} & $\times$ & - &  3.3B Words  & Collected \\
    \multirow{1}{*}{training} & 
    \multicolumn{1}{l|}{\multirow{1}{*}{LUPerson~\cite{fu2021unsupervised}}} & $\checkmark$ & 4M & - & - \\
    \cline{2-6}
    & \multicolumn{1}{l|}{\multirow{1}{*}{LUPerson-\textbf{T} (Ours)}} & $\checkmark$ & 1.3M & 1.3M & Generated \\
    \hline
    \multirow{3}{*}{T2I-ReID} & \multicolumn{1}{l|}{\multirow{1}{*}{CUHK-PEDES~\cite{li2017person}}} & $\checkmark$ & 40,206 & 80,412 & Labeled \\
     & \multicolumn{1}{l|}{\multirow{1}{*}{ICFG-PEDES~\cite{ding2021semantically}}} & $\checkmark$ & 54,522 & 54,522 & Labeled \\
     & \multicolumn{1}{l|}{\multirow{1}{*}{RESPReid~\cite{zhu2021dssl}}} & $\checkmark$ & 20,505 & 40,110 & Labeled \\
    \hline
% \end{tabular}
\end{tabu}}
\vspace{0.2cm}
\caption{The statistics comparison on the pre-training datasets and the T2I-ReID datasets. 
% $*$ means words.
}
\vspace{-0.2cm}
\label{tab:statistic}
\end{table}

\begin{figure*}[t!] % use float package if you want it here
  \centering
  \includegraphics[width=1.0\linewidth]{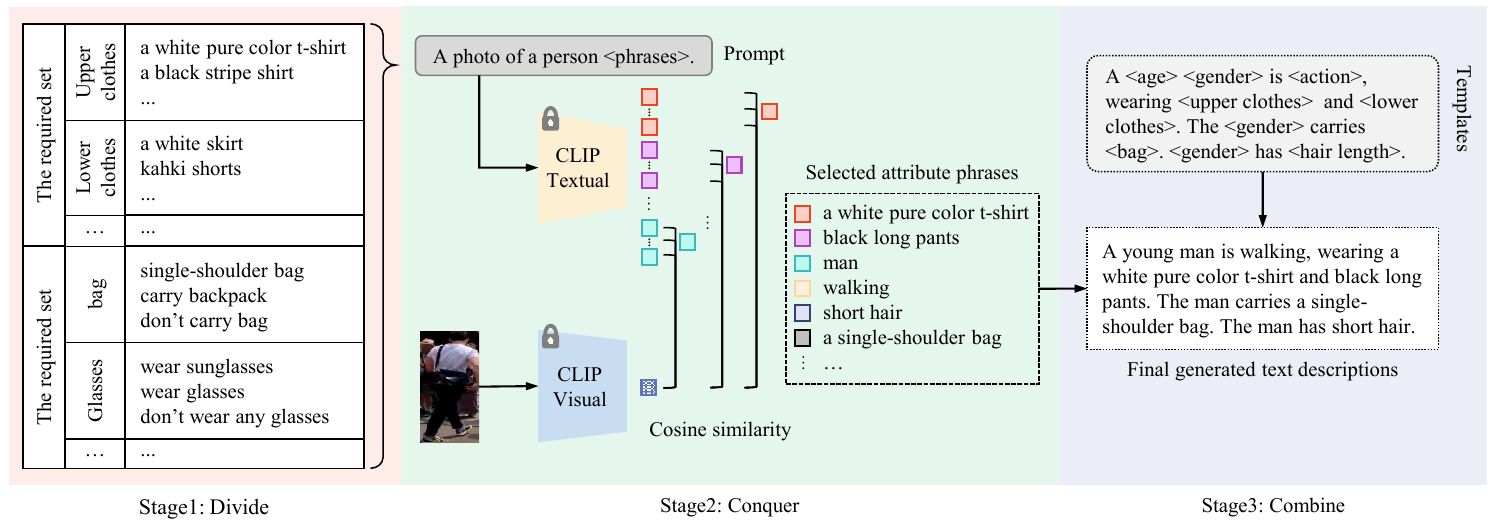}
  \caption{
  The overview of the pseudo text generation in LUPerson-T via the proposed divide-conquer-combine strategy. 
  In stage1, we \textit{divide} the person characters into attribute phrases.
  % the person characters into several attributes, and create one candidate phrase list for every atribute.
  In stage2, we assemble these attribute phrases into prompts, and send them to the frozen CLIP model.
  % to \textit{conquer} the subproblem,
  % we assemble these phrases with handcraft prompt and send them to frozen CLIP.
  According to the cosine similarity between text and image embeddings,
  we choose the most matched attribute phrases.
  In stage3, we \textit{combine} the matched attributes into the pre-defined templates to obtain complete sentences.
  % with the help of our handcraft template.
 }
\label{fig:lupersont}
\vspace{-0.2cm}
\end{figure*}

\section{Related Work}

\subsection{Vision-Language Pre-training Models}
The Transformer has attracted a lot of attention in the fields of computer vision and natural language processing. 
Many vision-language pre-training tasks apply the Transformer as their base structure~\cite{chen2020uniter,li2020unicoder,li2019visualbert,lu2019vilbert,su2019vl,sun2019videobert,tan2019lxmert,zhou2020unified}.
Depending on their model structure, the existing vision-language pre-trainin (VLPT) methods can be categorized into two-stream and single-stream models.
Several two-stream models \cite{su2019vl,sun2019videobert,tan2019lxmert} fuse two modalities after extracting visual features and textual features respectively.
Others, using an approach similar to CLIP, extract the features of two modalities completely separately and only employ contrastive loss at the end of the process.
For their part, the single-stream models \cite{chen2020uniter,li2020unicoder,li2019visualbert,lu2019vilbert,zhou2020unified} process the image feature and the language feature together as a joint distribution. 
Approaches of this kind require the paired text and image to be fed into the network together. 
% Specifically, in the test stage, these approaches brings in high computational complexities.
Recently, several works \cite{you2022learning,wang2022image} have attempted to explore the use of a shared backbone for both modalities. \cite{you2022learning} explores a Modality-Shared CLIP architecture, in which the visual encoder and textual encoder share the parameters. Besides, \cite{wang2022image} explores unified models for language, vision, and multimodal pre-training.

\subsection{Text-to-Image Person ReID}
Recently, text-to-image ReID has begun to attract more research attention, despite being more challenging than the general image-based ReID tasks~\cite{zheng2015scalable,sun2018beyond,zhang2019self,zhong2020learning,zhang2020memorizing,zheng2019joint,zhang2022implicit,fu2019horizontal,fu2022large,he2021transreid,luo2021self, yang2022unleashing,zhu2022pass} and the general cross-modal retrieval tasks. 
Existing works of this kind can be divided into global-embedding methods and local-embedding methods.
Methods in the former category \cite{zheng2020dual,zhang2018deep,wang2019language} directly create one dual-stream network to extract the features of different modalities, which align the global feature vectors to a common space.
Methods in the latter category \cite{li2017person,li2017identity,chen2018improving,jing2020pose,niu2020improving,wang2021text} design various kinds of model structures or objective functions to obtain and align the local features. For example, MGEL \cite{wang2021text} used a coarse-to-fine approach to learn local visual features on different spatial scales for different words or phrases. 
% SSAN \cite{ding2021semantically} used a self-aligned network instead of a cross-modal cross-attention module, which efficiently extracts semantically aligned visual and textual part features.
These local-embedding methods are usually based on a two-stream backbone, to which several local alignment modules are added. 
IVT \cite{shu2023see} utilized a single network to learn representation for both modalities. The network shares the same multi-head attention module but allots different FFNs to different modalities.
% IVT also employed two implicit semantic alignment paradigms, Multi-Level Alignment and Bidirectional Mask Modeling, to focus on more sementic and finer alignments. 
However, it faces one significant problem: pre-training. 
% At the textual side, the initial parameters of multi-head self-attention are from ViT, while the initial parameters of feed-forword module are from the BERT. They are not consistent.
% At the textual side, the initial parameters of multi-head self-attention which are from ViT and the initial parameters of feed-forword module which are from the BERT are not consistent. 
% Despite having pre-trained on four image captioning datasets \cite{radford2021learning,sharma2018conceptual,lin2014microsoft,krishna2017visual}, the network can not see pedestrian pictures in this stage. 
% LUPerson is a unlabeled ReID dataset which is collected from web videos. 
To address this issue, in this paper, we automatically generate natural language description tags for LUPerson's pedestrian images and utilize this pseudo-labeled dataset for our pre-training.

\section{LUPerson-T: LUPerson with Pseudo-Texts}
% A good pre-trained model can successfully increase the model's  generalization capability, based on large-scale
% % It usually relies on large-scale pre-training 
% datasets~\cite{russakovsky2015imagenet,devlin2018bert,xxx,xxx}.
Existing T2I-ReID datasets cover very limited data scales.
As shown in Table~\ref{tab:statistic}, the largest T2I-ReID dataset, ICFG-PEDES~\cite{ding2021semantically}, only contains fewer than $0.06$M image-text pairs.
Therefore, current T2I-ReID methods~\cite{zheng2020dual, zhang2018deep, wang2019language, liu2019deep,niu2020textual} directly use models pre-trained on large-scale databases like ImageNet~\cite{russakovsky2015imagenet} or generic texts~\cite{devlin2018bert}.
However, these approaches overlook the fact that a large domain gap exists in these contexts, since the pre-training data are inconsistent with the T2I-ReID data.
In detail, generic images consist of various objects and scenes that are not images of specific individual people, while T2I-ReID datasets only contain pedestrians.
Reviewing image-based ReID~\cite{fu2019horizontal,luo2021self,fu2022large}, pre-training on the large-scale unlabeled person dataset LUPerson~\cite{fu2021unsupervised} has demonstrated the effectiveness of pre-training on person images.
It is practicable to replace the initial visual encoder with a model pre-trained on LUPerson; however, the textual inconsistency still exists under these circumstances.
Moreover, the generic textual descriptions are usually coarse, while those in T2I-ReID datasets are fine-grained to describe each person’s characteristics.
We therefore naturally pose a question: \textit{can we improve the pre-training feature representations with a large-scale text-image person dataset for T2I-ReID?}

In this paper, we create a text-labeled variant of LUPerson~\cite{fu2021unsupervised}, named LUPerson-\textbf{T}.
LUPerson-T consists of 1.3M person images (a subset taken from~\cite{luo2021self}) along with their corresponding textual descriptions.
Dispensing with the need for manual annotations, 
we propose a \textit{divide-conquer-combine} strategy to generate pseudo-texts for person images.
In the following, we will carefully introduce this strategy.
To the best of our knowledge, LUPerson-T is the first visual-and-language pre-training dataset aimed at pedestrians. We make LUPerson-T public and hope that this dataset can help to progress research in the area of person ReID.

\begin{figure}[t]
\begin{center}
\includegraphics[width=0.95\linewidth]{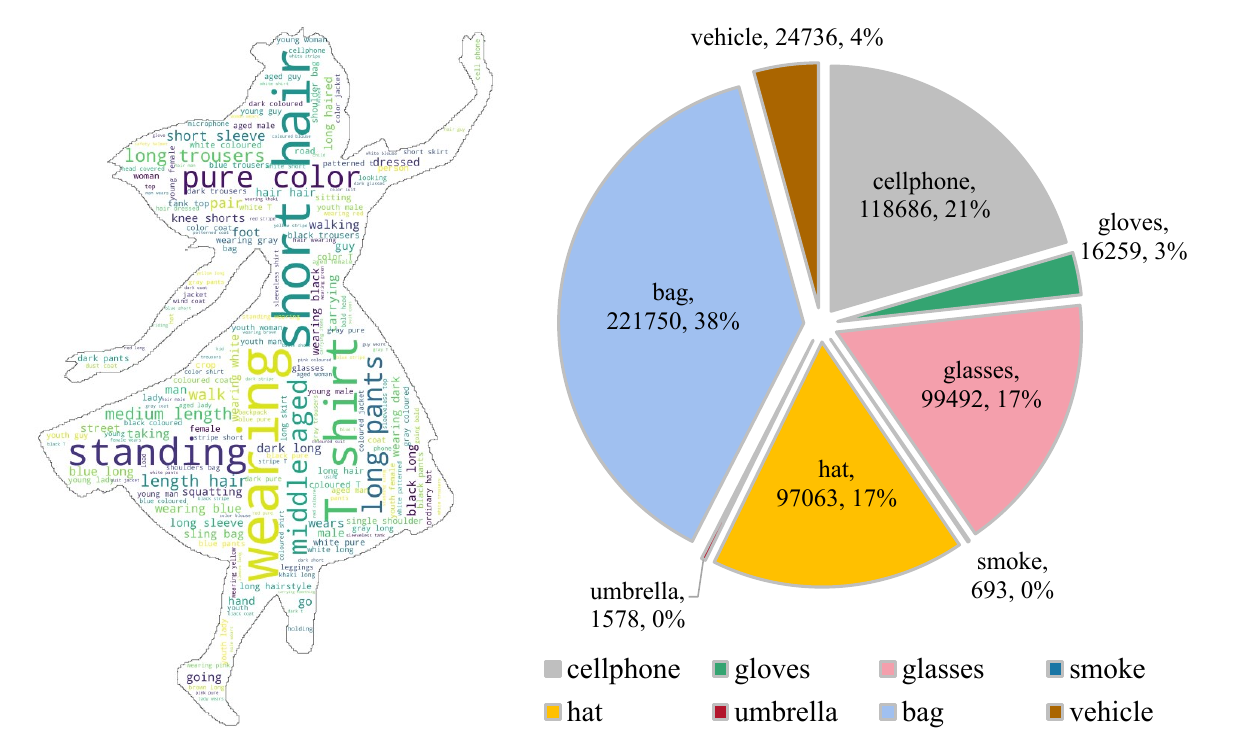}
\end{center}
  \caption{The WordCloud of high-frequency words (left) and the frequency of attributes in the optional set (right).}
\label{fig:fig_alice}
\vspace{-0.5cm}
\end{figure}

\subsection{Divide-Conquer-Combine Strategy: Pseudo Text Generation}
Since the manual annotation of person descriptions is time-consuming, we opt to automatically generate pseudo-texts.
Inspired by CLIP~\cite{radford2021learning}, a recent language-image pre-training approach that performs well on zero-shot tasks, we make an effort to obtain matched textual descriptions for a specific person image.
However, it is challenging for CLIP to directly generate complete and fine-grained person descriptions.
We thus propose a \textit{divide-conquer-combine} strategy to split a difficult problem into sub-problems, then integrate the corresponding sub-solutions into a complete one. 
The overview is presented in Figure~\ref{fig:lupersont}.

\myparagraph{Divide.} 
First of all, 
% we extract 14 kinds of attributes that have a large proportion appearing in the existing T2I-ReID datasets.
we examined most person images in the dataset and extract 14 kinds of attributes based on their frequency of occurrence.
% According to the frequency of occurrence, 
We intuitively group these attributes into two sets, \ie, the required set and the optional set.
The required set includes 6 attributes, namely {\textless age\textgreater}, {\textless gender\textgreater}, {\textless upper clothes\textgreater}, {\textless lower clothes\textgreater}, {\textless action\textgreater} and {\textless hair length\textgreater}, meaning that every person must possess all of these attributes\footnote{To be more precise, each person has a high probability of exhibiting each of the required attributes, since there may exist blur, occlusion, truncation, and so on in person images. For simplicity, we do not consider these problems in this paper.}. 
The optional set contains attributes that a person is likely to have, including {\textless bag\textgreater}, {\textless glasses\textgreater}, {\textless smoke\textgreater}, {\textless hat\textgreater}, {\textless cellphone\textgreater}, {\textless umbrella\textgreater}, {\textless gloves\textgreater} and {\textless vehicle\textgreater}, for a total of eight attributes.
These 14 attributes cover most basic aspects of pedestrian appearances and already facilitate good pre-training\footnote{We conjecture that introducing more attributes is potential to bring further improvement that we leave for the future work.}.
Each attribute additionally has its own phrases based on its classes.
For example, {\textless upper clothes\textgreater} contains ``a blue striped shirt'', ``a white pure color t-shirt'', ``a blue plaid
wind coat'' and so on.
Please refer to the supplementary material for more details.
These phrases are initial version in this stage that are artificially designed by ourselves, which will be then extended by synonym substitution in the combine stage.
% We first artificially design the initial phrases, then extend them by means of synonym substitution. Please refer to the supplementary material for more details.

\myparagraph{Conquer.} This is the most important step of the text generation. To make better use of the CLIP model, we convert attribute phrases into the standard prompt format, \ie, ``A photo of a person \textless phrases\textgreater''\footnote{There are minor modifications based on different phrases, \eg, ``A photo of a {\textless gender\textgreater}''.}.
We then put the person images into the frozen CLIP visual model to obtain visual representations, and the prompts into the frozen CLIP textual model to get textual features.
Here, we use the ViT-B/16 model \cite{openaiclip}. 
Subsequently, for each person image feature, we calculate cosine similarities between itself and all prompt features.
Then, for each of the required attributes, we choose the most similar phrase. For each of the optional attributes, we choose the satisfied phrase with a softmax probability larger than 0.9.
In this way, we can select the most relevant attribute phrases for person images.

\myparagraph{Combine.}
Next, the selected attribute phrases should be combined together to produce complete sentences.
In detail, we pre-defined 456 templates with ``blank'' positions for phrases, \eg,
``The {\textless age\textgreater} {\textless gender\textgreater} is {\textless action\textgreater}, 
wearing {\textless upper clothes\textgreater} and {\textless lower clothes\textgreater}. The {\textless gender\textgreater} wears {\textless glasses\textgreater} and carries {\textless bag\textgreater}. {\textless gender\textgreater} has {\textless hair length\textgreater}.''
Different templates have different patterns and diverse attribute blanks.
For a person image, we randomly choose one template, then automatically fill its candidate phrases from the conquer stage into the corresponding blanks.
It is common knowledge that the richness of vocabulary has a large impact on the quality of a dataset; we thus also use synonym substitution to enrich the diversity of the textual descriptions.

The proposed divide-conquer-combine strategy is automatic. Although we provide pre-defined templates, this effort is trivial compared to manual annotation. Based on this strategy, we can easily obtain a large-scale person-specific dataset with image and text pairs.

\subsection{Properties of LUPerson-T}
%数据集图片文本数量，模板数量，属性出现频率，
Table~\ref{tab:statistic} provides the statistics of LUPerson-T and existing T2I-ReID datasets up to now.
LUPerson-T contains about 1.3M image–text person pairs, making it the largest T2I-ReID dataset in existence, even though the text descriptions are pseudo-texts. 
% There are a total of 26,100,063 words and 8,111 unique words in the dataset.
The optional set appears 580,257 times in total, in which the occurrence frequency of four attributes accounts for 94\% of all optional attributes (namely {\textless bag\textgreater}, {\textless cellphone\textgreater}, {\textless glasses\textgreater} and {\textless hat\textgreater}).
Among them, the {\textless bag\textgreater} attribute appears most frequently (about a 38\% chance).

% \zhiyin{
% Currently, the quantity of all sentence templates is 456 and there are approximately 20.4 words for each sentence on average.
% There are a total of 26,100,063 words and 8,111 unique words in
% our dataset. 
% The variable attribute appears 580,257 times in total and the three variable attributes that occur the most are bag, cellphone, glasses.
% Figure~\ref{fig:fig_alice} illustrates the frequency of variable attributes and some high-frequency words.
% }

% Although LUPerson-T is built based on the LUPerson, our method to generate pseudo-texts theoretically can be applied to other pedestrian datasets.
% The fixed attributes mentioned in \ref{sec:setion3.1} include the upper\_clothes, lower\_clothes, action, hair\_length, age and gender.
% The phrase of upper\_clothes and lower\_clothes also can be devided into three tags: color, texture and the type of clothes.
% The type of upper\_clothes include 
% Currently, the quantity of all description templates is 456 and there are approximately 19.3 words for each description on average.
% By default, original LUPerson contains 4,180,243 images and we create at least two descriptions for each image.
% We set nine attribute categories as fixed attributes, and the rest of nine attribute categories as variable attributes.

\section{Unified Pre-Training for T2I-ReID Task}
Based on LUPerson-T, we can apply vision-and-language pre-training frameworks.
Unlike solely visual or solely textual pre-training, these frameworks can perform interactive learning between images and texts.
In other words, the pre-training shares a unified behavior as T2I-ReID tasks.
For simplicity, we refer to this unified pre-training pipeline on LUPerson-T as \textbf{UniPT}.
% In the following, we will introduce the training details of the pre-training on LUPerson-T and the supervised learning on T2I-ReID datasets. 

%先说基于LUperson-T，我们能够使用VLP的方式进行pretrain了。因此，pretrain和下游的训练方式就consistent了。因此，我们称这个pretrain是unified

\begin{figure}[t!]
\begin{center}
\includegraphics[width=0.95\linewidth]{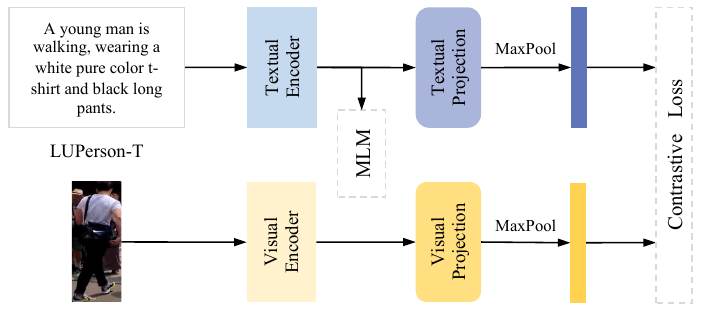}
\end{center}
\vspace{-0.2cm}
  \caption{Overview of our UniPT pipeline.}
\label{fig:fig_arch}
\vspace{-0.3cm}
\end{figure}

\subsection{UniPT: Unified Pre-training on LUPerson-T}
We follow CLIP~\cite{radford2021learning} as our basic framework.
%\footnote{We acknowledge that other well-designed vision-and-language pre-training frameworks may be superior than CLIP, however, it is not the main concern in this paper. We leave this exploration in the future work.}.
We feed images and texts into the image encoder and text encoder to obtain visual and textual features, respectively.
These two modality features are forced into a multi-modal embedding space by jointly training the visual and textual encoders.
The objective function is the contrastive loss~\cite{zhang2022contrastive, oord2018representation,radford2021learning} on visual and textual feature representations.
% It is because the generic image-text pre-training gains benefit from the training consistency with T2I-ReID, i.e., both tasks align the visual and textual features.
Besides, we also introduce the mask language model (MLM) pre-training objective in \cite{devlin2018bert} to prevent the underlying over-fitting.

\myparagraph{Contrastive loss on image-text pairs.}
Given a mini-batch of $N$ image-text pairs, we aim to predict which of the $N \times N$ possible image-text pairings across a batch are truly matched (or not).
The cosine similarity matrix of $N \times N$ image-text pairings is calculated on visual embeddings ${\pmb{V}}=\{ {\pmb{v}}_1, {\pmb{v}}_2, ..., {\pmb{v}}_N\}$ and textual embeddings ${\pmb{T}}=\{ {\pmb{t}}_1, {\pmb{t}}_2, ..., {\pmb{t}}_N\}$.
The contrastive loss acts to maximize the cosine similarity of the  real matched pairs of image and text embeddings while minimizing that of the $N^2$ - $N$ incorrect pairings.
We apply the symmetric contrastive loss function used in \cite{radford2021learning}, which is formulated as follows:
\begin{equation}
\begin{aligned}
% L_{con} = {\dfrac{1}{N}} \sum_{n=1}^N -log ({\dfrac{exp({\bf{v}}_{n} \cdot {\bf{t}}_{+})}{\sum_{i=1}^N exp({\bf{v}}_{n} \cdot {\bf{t}}_{i})}})\\
% + {\dfrac{1}{N}} \sum_{n=1}^N -log ({\dfrac{exp({\bf{t}}_{n} \cdot {\bf{v}}_{+})}{\sum_{i=1}^N exp({\bf{t}}_{n} \cdot {\bf{v}}_{i})}}),
L_{\mathrm{I2T}} = {\dfrac{1}{N}} \sum_{n=1}^N & -{\mathrm{log}} ({\dfrac{{\mathrm{exp}}({\mathrm{sim}}({\pmb{v}}_{n}, {\pmb{t}}_{+}))}{\sum_{i=1}^N {\mathrm{exp}}({\mathrm{sim}}({\pmb{v}}_{n},{\pmb{t}}_{i}))}}),\\
L_{\mathrm{T2I}} = {\dfrac{1}{N}} \sum_{n=1}^N & -{\mathrm{log}} ({\dfrac{{\mathrm{exp}}({\mathrm{sim}}({\pmb{t}}_{n}, {\pmb{v}}_{+}))}{\sum_{i=1}^N {\mathrm{exp}}({\mathrm{sim}}({\pmb{t}}_{n}, {\pmb{v}}_{i}))}}),\\
L_{\mathrm{con}} & = (L_{\mathrm{I2T}} + L_{\mathrm{T2I}}) / 2,
\label{eq:con_loss}
\end{aligned}
\end{equation}
where ${\pmb{v}}_{i}$/${\pmb{t}}_{i}$ is the $i$-th image/text representation of a batch.
${\pmb{t}}_{+}$ is the corresponding textual description of ${\pmb{v}}_{n}$.
Similarly, ${\pmb{v}}_{+}$ and ${\pmb{t}}_{n}$ are a matched image-text pair.  
$\mathrm{sim}(\pmb{a}, \pmb{b})=\pmb{a}^{\mathrm{T}}\pmb{b}/ \left|\left| \pmb{a} \right|\right|\left|\left| \pmb{b} \right|\right|$ denotes the cosine similarity between the two vectors $\pmb{a}$ and $\pmb{b}$.

% \begin{equation}
% S({\bf{v}},{\bf{t}}) = {\dfrac{{\bf{v}}^{T}{\bf{t}}}{\Vert {\bf{v}} \Vert \Vert {\bf{t}} \Vert}},
% \end{equation}
% Here, ${\bf{t}}_{n}$/${\bf{v}}_{n}$ are the query representation,
% ${\bf{v}}_{+}$/${\bf{t}}_{+}$ are the representation of the positive sample.

% Although each image in LUPerson-T has its own textual pedestrian, there is no ID information.
% Inspired by recent visual-language pre-training works \cite{radford2021learning}, we also introduce the contrastive loss to train our architecture.

% We input the image and the text independently to the visual and textual encoder to extract high-level feature representations, using the global visual and textual features ({\bf{v}} and {\bf{t}}) to calculate contrastive loss between two modalities.
% Specifically, given a batch of N pairs of image and text, we could get predictions of $N \times N$ possible pairings, which are calculated by the the cosine similarity.
% The contrastive loss is to maximize the cosine similarity of the image and text embeddings of the $N$ real pairs in the batch while minimizing the cosine similarity of the embeddings of the $N^2$ - $N$ incorrect pairings.
% Formally,

\myparagraph{Mask language model objective on texts.}
Recall that CLIP contains 400M image–text pairs, representing adequate textual diversity. 
In contrast, the text descriptions in our LUPerson-T are specific for each person, constructed by a relatively small scale of attribute phrases.
Therefore, using only the contrastive loss may result in over-fitting.
Accordingly, in order to learn more robust representations, we also employ the mask language model (MLM) \cite{devlin2018bert} objective in our pre-training.
%%% 为什么我们要用这个loss呢
% In order to learn more robust representations, we employ Mask Language Model (MLM) \cite{devlin2018bert} in our pre-training.
Specifically, we randomly mask 15\% of the tokens in a sentence, following \cite{devlin2018bert}.
The objective of MLM is to predict 
% the original vocabulary id (ground truth) of 
the masked words based on their context.
Let $x$ denote a person’s textual description, while $\mathcal{M}$ is the set of the randomly masked positions of $x$.
The formula of the MLM loss is defined as follows:
% And the final hidden vectors corresponding to the mask token are used to predict the ground truth.
% Formally,
\begin{equation}
\label{eq:mlm}
L_{\mathrm{mlm}}=-\sum\limits_{i\in \mathcal{M}} {\mathrm{log}} ~p(x_i|x_{\backslash\mathcal{M}}),
%\mathop{\mathbb{E}}\limits_{x\in \mathcal{C}} \mathop{\mathbb{E}}\limits_{\substack{\mathcal{M}\subset x\\|\mathcal{M}|=m|x|}}\left[\sum\limits_{x_i\in \mathcal{M}} \log p(x_i|\tilde{x})\right],
\end{equation}
where $x_{\backslash\mathcal{M}}$ is the masked version of $x$.
The target of the masked word is its corresponding vocabulary id \cite{devlin2018bert}.
With MLM loss, the optimization of the pre-training task is made more difficult to prevent the over-fitting problem.

% where $\mathcal{M}$ denotes the masked token set, $\tilde{x}$ denotes the masked version of $x$. We mask $m$ percentage of tokens and predicts them on $\mathcal{M}$ given the corrupted context.

\myparagraph{Total optimization.}
We use the contrastive loss and the MLM loss together to optimize the overall UniPT framework.
% We input two kinds of versions of texts simultaneously, including the masked version and the intact version.
% The former one is for the MLM loss, while the latter one is used for the contrastive loss.
% We try to use the same masked texts for both two loss functions, however, the model is hard to converge to an optimal status (\xinyu{poor downstream performance as shown in Section xxx}).
The final loss function is as follows:
\begin{equation}
% \vspace{-0.1cm}
\centering
\begin{aligned}
L_{\mathrm{pre}}=L_{\mathrm{con}}+ \beta L_{\mathrm{mlm}},
\end{aligned}
% \vspace{-0.1cm}
\label{eq:overall_loss}
\end{equation} 
where $\beta=\{0, 1\}$ controls whether or not the MLM loss is used. 
When $\beta=0$, the optimization is reduced to using only contrastive loss.
By default, we set $\beta=1$ so that $L_{\mathrm{con}}$ and $L_{\mathrm{mlm}}$ are used together.
% (our default setting).
% is the loss weight, balancing two loss functions $L_{\mathrm{con}}$ and $L_{\mathrm{mlm}}$.
% By default, we set $\beta=1$.

\myparagraph{Model architecture.}
The model architecture is illustrated in Figure~\ref{fig:fig_arch}.
We attach a $1 \times 1$ convolution layer to the visual and textual encoder respectively to project the image and text features into the same dimension.
Subsequently, we conduct a max-pooling operation on both the visual and textual features to obtain the global embeddings of ${\pmb{V}}$ and ${\pmb{T}}$ for model training.

\subsection{Supervised Learning on T2I-ReID datasets}\label{sec:setion4.3}

% \section{Architecture and Training}
% Our architecture and pre-training strategy is illustrated in \ref{fig:lupersont}(right).
% First, we introduce the pre-training strategy and architecture in \ref{sec:setion4.1}.
% The training in the downstream stage are described in \ref{sec:setion4.2}.
% \subsection{Learning in Downstream Stage}\label{sec:setion4.3}
Based on our UniPT models, we next
conduct supervised fine-tuning on the downstream T2I-ReID datasets.
% with identification labels.
% As the same as the pre-training stage, we extract global textual and visual embeddings independently, and then feed then into 
We use the identification loss \cite{sun2018beyond,zhang2018deep} and ranking loss \cite{faghri2017vse++} on global and granularity-unified features \cite{shao2022learning} for optimization.
% Besides, we also employ the prototype-based granularity unification (PGU) module in \cite{shao2022learning} to extract the granularity-unified features for better optimization with the ranking loss.

\myparagraph{Identification loss.} Inspired by \cite{sun2018beyond,zhang2018deep}, we adopt cross-entropy loss as the identification (ID) loss.
For a specific visual feature ${\pmb{v}}$ and textual feature ${\pmb{t}}$, we denote the predicted identity probability as $\hat{{y}}_{v}$ and $\hat{{y}}_{t}$, respectively.
% In the same way, the predicted identity probabilities is  $\hat{{\pmb{y}}}_{t}$ for the textual feature ${\pmb{t}}$.
The identification loss is formulated as follows:
% \begin{equation}
% L_{ID}({\bf{v}}) = {\dfrac{1}{K}} \sum_{k=1}^K -{\bf{y}}~{\odot}~log (\hat{{\bf{y}}}^{v}_k),
% \label{eq:id_loss}
% \end{equation}
% \begin{equation}
% L_{ID}({\bf{t}}) = {\dfrac{1}{K}} \sum_{k=1}^K -{\bf{y}}~{\odot}~log (\hat{{\bf{y}}}^{t}_k),
% \label{eq:id_loss}
% \end{equation}
\begin{equation}
\begin{aligned}
L_{\mathrm{id}} ({\pmb{v}}, {\pmb{t}})=-({y}~\mathrm{log}~\hat{{y}}_{v} + {y}~\mathrm{log}~\hat{{y}}_{t}), 
% L_{\mathrm{id}}({\pmb{v}},{\pmb{t}}) 
% &={\dfrac{1}{K}} \sum_{k=1}^K -{\bf{y}}~{\odot}~log (\hat{{\bf{y}}}^{v}_k)\\
% &+{\dfrac{1}{K}} \sum_{k=1}^K -{\bf{y}}~{\odot}~log (\hat{{\bf{y}}}^{t}_k),
\end{aligned}
\label{eq:id_loss}
\end{equation}
% where $\hat{{\pmb{y}}}_{v}$ and $\hat{{\pmb{y}}}_{t}$ denote the predicted identity probabilities of visual and textual features.
where ${y}$ is the ground-truth person ID label for the ${\pmb{v}}$ and ${\pmb{t}}$ pair. 
Following \cite{ding2021semantically}, we use a shared classifier to get image and text probabilities.
% Here, to further narrow the inter-modality distance, global visual and textual features share one classifier which output the predicted identity probabilities $\hat{\bf{y}}_{k}$. The identification loss is show as:

\begin{table}[t!]
\centering
\small
\setlength{\tabcolsep}{0.2mm}{
 % to 0.9\linewidth {l|X[c]|X[c]|X[c]|X[c]|X[c]|X[c]|X[c]}
\begin{tabu} to 0.7\linewidth {c|c|c|cc|cc}
    \hline
        \multicolumn{2}{c|}{Pre-training data} & 
        \multicolumn{1}{c|}{\multirow{2}{*}{Backbone}} &
        \multicolumn{2}{c|}{CUHK-PEDES} & \multicolumn{2}{c}{ICFG-PEDES} \\
    \cline{1-2}
    \cline{4-7}
    % \hline
        Image & Text & & Rank-1 & Rank-5 & Rank-1 & Rank-5\\
    % \hline
      
    \hline\hline
    \multicolumn{1}{c|}{\multirow{2}{*}{IN~\cite{russakovsky2015imagenet}}}
        & \multicolumn{1}{c|}{\multirow{2}{*}{Text~\cite{devlin2018bert}}} & DeiT-small \cite{touvron2021training} & 65.30 & 83.19 & 57.60 & 74.53 \\
       & & ViT-B/16 \cite{dosovitskiy2020image} & 66.59 & 84.40 & 58.62 & 75.31 \\
       % \hline
       % CLIP (Res50) \cite{radford2021learning} & CLIP~\cite{radford2021learning} & 61.25 & 81.34 &  52.34 & 71.19 \\
       % CLIP (ViT-B/16) \cite{radford2021learning} & CLIP~\cite{radford2021learning} & 66.34 & 84.18 & 59.01 & 75.96\\
       % ViLT-B/32 \cite{kim2021vilt} & - & 54.60 & 74.87 & 83.33 & 51.16 & 49.53& 67.23 & 75.31 & 27.13\\
       % DeiT-small \cite{touvron2021training} & - & 60.56 & 79.09 & 85.61 & 52.21 & 51.66& 69.06& 76.36& 28.68 \\
       \hline
       \multicolumn{1}{c|}{\multirow{2}{*}{LUP-T}} & \multicolumn{1}{c|}{\multirow{2}{*}{LUP-T}} & DeiT-small~\cite{touvron2021training} & 66.83 & 84.16 & 59.08 & 75.92\\
       & & ViT-B/16 \cite{dosovitskiy2020image} & \textbf{68.50} & \textbf{84.67}  & \textbf{60.09} & \textbf{76.19}\\
       %t2i: @R1: 0.6683, @R5: 0.8416, @R10: 0.8942, map: 0.5816
         %t2i: @R1: 0.5908, @R5: 0.7592, @R10: 0.8208, map: 0.3399
    \hline
\end{tabu}}
\vspace{0.2cm}
\caption{LUPerson-T boosts the T2I-ReID performance.
Previous works~\cite{zheng2020dual, zhang2018deep, wang2019language, liu2019deep,niu2020textual} use the pre-trained models on ImageNet~\cite{russakovsky2015imagenet} and generic texts~\cite{devlin2018bert} as the visual and textual models (in the top block).
Differently, our UniPT utilizes the pre-trained models on the proposed LUPerson-T (in the bottom block).
% We evaluate on both DeiT-small~\cite{touvron2021training} and ViT-B/16 \cite{dosovitskiy2020image} backbones.
The abbreviations \{``IN'', 'Text', 'LUP-T'\} represent ImageNet~\cite{russakovsky2015imagenet}, generic texts in \cite{devlin2018bert} and our LUPerson-T respectively.
}
\label{tab:luperson_t}
\end{table}

\begin{table}[t!]
\centering
\small
\setlength{\tabcolsep}{0.8mm}{
 % to 0.9\linewidth {l|X[c]|X[c]|X[c]|X[c]|X[c]|X[c]|X[c]}
\begin{tabu} to 0.95\linewidth {l|c|c|cc|cc}
\hline
    \multicolumn{1}{l|}{\multirow{2}{*}{Type}} & \multicolumn{2}{c|}{Pre-training models} & 
    \multicolumn{2}{c|}{CUHK-PEDES} & \multicolumn{2}{c}{ICFG-PEDES} \\
    \cline{2-7}
      & Visual & Textual   & Rank-1  & Rank-5 & Rank-1  & Rank-5\\
    \hline
    \hline
    %& 60.14  &77.33 & 47.25 & 66.58 \\
    Single & SSL~\cite{luo2021self} & BERT~\cite{devlin2018bert} & 60.14  &77.33 & 47.25 & 66.58 \\
    Single & Ours & BERT~\cite{devlin2018bert} & 65.01  & 82.60 & 58.15 & 75.05 \\
    UniPT & Ours & Ours & \textbf{66.83} & \textbf{84.16} & \textbf{59.08} & \textbf{75.92} \\
    \hline
\end{tabu}}
\vspace{0.2cm}
\caption{Our unified pre-training pipeline (UniPT) improves the feature representations of both visual and textual modalities. 
``Single'' means that the visual and textual models are pre-trained independently.
For a fair comparison, all experiments use small-size ViT as the visual backbone, and BERT as the textual backbone.}
\vspace{-0.2cm}
\label{tab:pretrain}
\end{table}

\myparagraph{Ranking loss.}
The ranking loss \cite{faghri2017vse++} is commonly used to align visual and textual modalities.
It is based on the similarity of the text-image triplets, which is formulated as follows:
\begin{equation}
\begin{aligned}
L_{\mathrm{rk}} ({\pmb{v}}, {\pmb{t}}) = &~\mathrm{max}(\alpha - \mathrm{sim}({\pmb{v}}, {\pmb{t}}_{+}) + \mathrm{sim}({\pmb{v}}, {\bf{t}}_{-}), 0) \\
+ &~\mathrm{max}(\alpha - \mathrm{sim}({\pmb{t}}, {\pmb{v}}_{+}) + \mathrm{sim}({\pmb{t}}, {\pmb{v}}_{-}), 0),
% L_{\mathrm{rk}}({\bf{v}},{\bf{t}}) 
% &= max(\alpha - S({\bf{v}}, {\bf{t}}^{+}) + S({\bf{v}}, {\bf{t}}^{-}),0)\\
% &+max(\alpha - S({\bf{t}}, {\bf{v}}^{+}) + S({\bf{t}}, {\bf{v}}^{-}),0),
\end{aligned}
\label{eq:tri_loss}
\end{equation}
where ${\pmb{t}}_{+}$/${\pmb{v}}_{+}$ and ${\pmb{t}}_{-}$/${\pmb{v}}_{-}$ denote the positive sample and the semi-hard negative sample~\cite{ge2019visual} of ${\pmb{v}}$/${\pmb{t}}$ respectively.
$\alpha$ is the hyper-parameter of the margin.

\myparagraph{Granularity-unified loss.}
In the T2I-ReID task, the granularity alignment is key to reducing the modality gap between text and image~\cite{shao2022learning}.
Therefore, we also employ the prototype-based granularity unification (PGU) module presented in \cite{shao2022learning} to extract the granularity-unified features 
% $\pmb{\widetilde{v}}$ and $\pmb{\widetilde{t}}$ 
$\widetilde{\pmb{v}}$ and $\widetilde{\pmb{t}}$.
Following \cite{shao2022learning}, we apply the identification loss and the ranking loss on $\widetilde{\pmb{v}}$ and $\widetilde{\pmb{t}}$, which is formulated as follows:
\begin{equation}
\begin{aligned}
L_{\mathrm{pgu}} = L_{\mathrm{id}} (\widetilde{\pmb{v}}, \widetilde{\pmb{t}}) + L_{\mathrm{rk}}(\widetilde{\pmb{v}}, \widetilde{\pmb{t}}).
\end{aligned}
\label{eq:pgu_loss}
\end{equation}

\myparagraph{Total optimization.}
The overall loss function in the T2I-ReID supervised fine-tuning stage is as follows:
% The global loss can be denoted as follows:
\begin{equation}
\begin{aligned}
L_{\mathrm{ft}} = L_{\mathrm{id}} + L_{\mathrm{rk}} + \gamma L_{\mathrm{pgu}}
% L_{\mathrm{g}} = L_{ID}({\bf{v}}_{g},{\bf{t}}_{g}) + L_{RK}({\bf{v}}_{g},{\bf{t}}_{g}),
\end{aligned}
\label{eq:ft_loss}
\end{equation}
where $\gamma=\{0, 1\}$ controls whether or not the granularity-unified loss is used. 
When $\gamma=0$, we only employ $L_{\mathrm{id}}$ and $L_{\mathrm{rk}}$ on $\pmb{v}$ and $\pmb{t}$: when $\gamma=1$, the additional $L_{\mathrm{pgu}}$ is added for the supervised learning (our default setting).
\begin{figure}[t!]
\begin{center}
\includegraphics[width=0.95\linewidth]{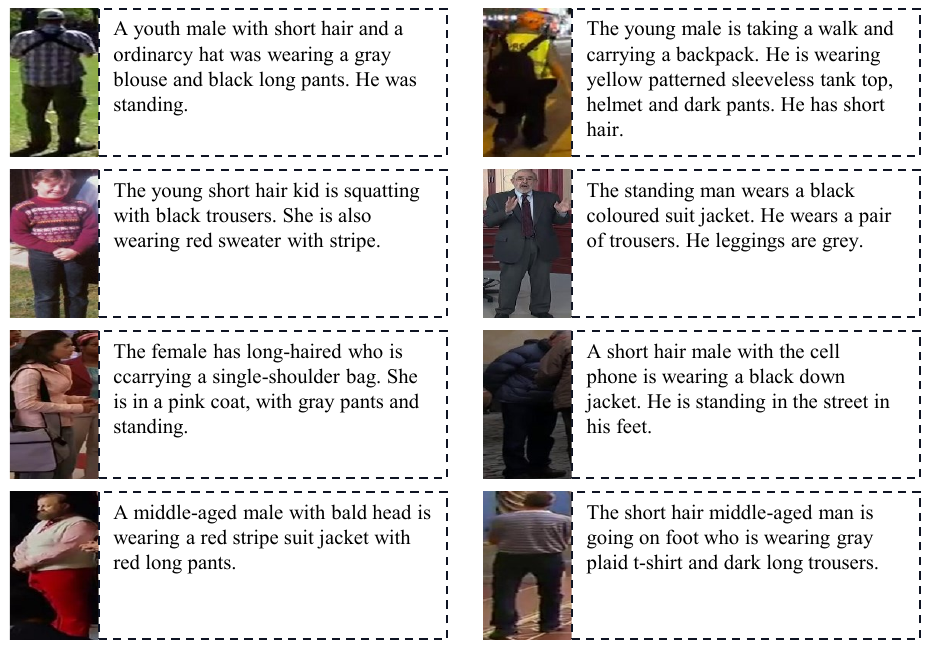}
\end{center}
\vspace{-0.3cm}
  \caption{Visualization of some examples in LUPerson-T. These sentences can fairly accurately describe the appearance of the human body in a fluent sentence pattern.}
\label{fig:visualization}
\vspace{-0.2cm}
\end{figure}

\begin{table}[t!]
\centering
\small
\setlength{\tabcolsep}{0.8mm}{
 % to 0.9\linewidth {l|X[c]|X[c]|X[c]|X[c]|X[c]|X[c]|X[c]}
\begin{tabu} to 0.7\linewidth {l|c|cc|cc}
    \hline
        \multicolumn{1}{l|}{\multirow{2}{*}{Method}} & 
        \multicolumn{1}{c|}{\multirow{2}{*}{Backbone}} &
        \multicolumn{2}{c|}{CUHK-PEDES} & \multicolumn{2}{c}{ICFG-PEDES} \\
    \cline{3-6}
    % \hline
         &  & Rank-1 & Rank-5 & Rank-1 & Rank-5\\
    % \hline
      
    \hline\hline
       % DeiT-small \cite{touvron2021training} & BERT~\cite{devlin2018bert} & 65.30 & 83.19 & 57.60& 74.53 \\
       % ViT-B/16 \cite{dosovitskiy2020image} &  BERT~\cite{devlin2018bert} & 66.59 & 84.40 & 58.62 & 75.31 \\
       % \hline
       \multicolumn{1}{l|}{\multirow{2}{*}{CLIP~\cite{radford2021learning}}} & ResNet50 & 61.25 & 81.34 & 52.34 & 71.19 \\
       & ViT-B/16 & 66.34 & 84.18 & 59.01 & 75.96\\
       % ViLT-B/32 \cite{kim2021vilt} & - & 54.60 & 74.87 & 83.33 & 51.16 & 49.53& 67.23 & 75.31 & 27.13\\
       % DeiT-small \cite{touvron2021training} & - & 60.56 & 79.09 & 85.61 & 52.21 & 51.66& 69.06& 76.36& 28.68 \\
       \hline
       \multicolumn{1}{l|}{\multirow{2}{*}{UniPT}} & DeiT-small & 66.83 & 84.16 & 59.08 & 75.92 \\
       & ViT-B/16 & \textbf{68.50} & \textbf{84.67}  & \textbf{60.09} & \textbf{76.19} \\
       %t2i: @R1: 0.6683, @R5: 0.8416, @R10: 0.8942, map: 0.5816
         %t2i: @R1: 0.5908, @R5: 0.7592, @R10: 0.8208, map: 0.3399
    \hline
\end{tabu}}
\vspace{0.2cm}
\caption{Our UniPT \vs CLIP~\cite{radford2021learning}.  
CLIP is also a vision-and-language pre-training framework.
Our UniPT pre-train on LUPerson-T, and CLIP uses the generic image-text pairs.
% Different from our UniPT on LUPerson-T, CLIP uses 400M private generic image-text pairs.
}
\vspace{-0.4cm}
\label{tab:ablation2}
\end{table}

\section{Experiments}
\subsection{Datasets and Evaluation Metrics}
\myparagraph{Datasets.} We evaluate three T2I-ReID datasets, \ie, CUHK-PEDES~\cite{li2017person}, ICFG-PEDES~\cite{ding2021semantically}, and RSTPReid~\cite{zhu2021dssl}. 
The statistics of these three datasets are listed in Table~\ref{tab:statistic}.

% \myparagraph{CUHK-PEDES} \cite{li2017person} is a large person search dataset which contains 40,206 images and 80,412 textual descriptions for 13,003 identities. 
% The training set comprises 34,050 images and 68,120 textual descriptions of 11,000 pedestrains; the validation set is made up of 3,078 images, 1,000 pedestrains and 6158 textual descriptions; finally, the test set includes 3,074 images and 6,156 textual descriptions of another 1,000 pedestrains. Each picture contains at least two textual descriptions, and there are on average 23.5 words for each textual description.

% \myparagraph{ICFG-PEDES} \cite{ding2021semantically} contains 54,522 pedestrian images of 4,102 different identities. 
% All pictures are collected from the MSMT17 database \cite{wei2018person}. The training set has 34,674 images, 3102 pedestrains and 34,674 textual descriptions.  The test set has 19,848 images, 1,000 pedestrians and 19,848 textual descriptions.
% Each picture contains only one textual descriptions, and there are on average 37.2 words for each textual description. 

% \myparagraph{RSTPReid} \cite{zhu2021dssl} contains 20,505 pedestrian images of 4,101 persons. 
% The training set, the validation set and the test set have 3,701, 2,00 and 2,00 identities respectively.
% Each identity has 5 images by different cameras and each image has 2 textual descriptions.

\myparagraph{Evaluation metrics.} We adopt the popular Rank-$k$ metrics ($k=1, 5, 10$) as the evaluation metrics. 
Rank-$k$ reveals the probability that when given a textual description as a query, at least one matching person image is found in the top-$k$ candidate list.

\subsection{Implementation Details}

\myparagraph{Model structure.}
In our experiments, we opt to use DeiT-Small~\cite{touvron2021training} and ViT-B/16~\cite{dosovitskiy2020image} as the visual backbones and BERT~\cite{devlin2018bert} as the textual backbone, respectively.

\myparagraph{Pre-training.}
The input image resolution is set to 384$\times$128 and the text token length is 100.
For all experiments, we use the AdamW~\cite{loshchilov2017decoupled} optimizer with a base learning rate of 1e-5. The learning rate is warmed up for 10\% of the total steps and the batch size is set to 512.
The total number of epochs is 15.
The model is pre-trained on 8 Nvidia Tesla V100 GPUs.
% \xinyu{image size, text token length, }
% \xinyu{add training details, like lr, batchsize, optimizer}
% The pre-training is conducted on 8 Nvidia Tesla V100 GPUs.

\myparagraph{Supervised learning.}
All person images are resized to 384$\times$128 and augmented with random horizontal flipping.
% The feature dimension $C$ for both the image and text is 768.
The margin $\alpha$ of the ranking loss is set to 0.2 in both $L_{\mathrm{rk}}$ and $L_{\mathrm{pgu}}$.
The batch size is 64, and the optimizer used is Adam~\cite{kingma2014adam}.
Except for the PGU module, with a learning rate of 1e-4, the initial learning rate is set to 1e-5 for the pre-trained textual encoder and 1e-4 for the visual encoder.
The training is conducted on a single Nvidia Tesla V100 GPU.

\begin{table}[t!]
\centering
\small
\setlength{\tabcolsep}{0.8mm}{
 % to 0.9\linewidth {l|X[c]|X[c]|X[c]|X[c]|X[c]|X[c]|X[c]}
\begin{tabu} to 0.9\linewidth {l|cc|cc}
\hline
    \multicolumn{1}{l|}{\multirow{2}{*}{Method}} & \multicolumn{2}{c|}{CUHK-PEDES} & \multicolumn{2}{c}{ICFG-PEDES} \\
    \cline{2-5}
      & Rank-1  & Rank-5 & Rank-1  & Rank-5\\
    \hline
    \hline
        UniPT w/o MLM  & 65.77  &83.53 &57.35 &74.14 \\
        UniPT w MLM (UniPT) & \textbf{66.83} & \textbf{84.16} & \textbf{59.08} & \textbf{75.92} \\
         %t2i: @R1: 0.6577, @R5: 0.8353, @R10: 0.8897, map: 0.5748
         %t2i: @R1: 0.5735, @R5: 0.7414, @R10: 0.8064, map: 0.3294
         
    \hline
\end{tabu}}
\vspace{0.2cm}
\caption{
Effectiveness of the masked language modeling (MLM) objective in our UniPT.
% Ablation study on loss weight $\beta$ in Eq.~\eqref{eq:overall_loss} in our unified pre-training. When $\beta=1$, the pre-training uses both contrastive loss and masked language model (MLM) objective; when $\beta=0$, we only use contrastive loss.
% We set $\beta$ to 1 by default.
The visual backbone is DeiT-small \cite{touvron2021training}.
% (marked in \colorbox{baselinecolor}{gray}).
}
% \vspace{-0.1cm}
\label{tab:mlm}
\end{table}

\begin{table}[t!]
\centering
\small
\setlength{\tabcolsep}{0.8mm}{
 % to 0.9\linewidth {l|X[c]|X[c]|X[c]|X[c]|X[c]|X[c]|X[c]}
\begin{tabu} to 0.7\linewidth {c|c|cc|cc}
    \hline
    \multicolumn{1}{l|}{\multirow{1}{*}{Pre-training}} & \multicolumn{1}{c|}{\multirow{2}{*}{Method}} & \multicolumn{2}{c|}{CUHK-PEDES} & \multicolumn{2}{c}{ICFG-PEDES} \\
    \cline{3-6}
    \multicolumn{1}{l|}{\multirow{1}{*}{Type}} & & Rank-1  & Rank-5 & Rank-1  & Rank-5\\
    \hline
    \hline
         % $\gamma=0$ & 60.56  &79.09 &51.66 &69.06\\
      \multicolumn{1}{l|}{\multirow{2}{*}{Single}} & w/o $L_{\mathrm{pgu}}$ & 60.56  &79.09 &51.66 &69.06\\
       & w/ $L_{\mathrm{pgu}}$ & 65.30  &83.19 &57.30 &74.18\\
       \hline 
      \multicolumn{1}{l|}{\multirow{2}{*}{UniPT}} &   w/o $L_{\mathrm{pgu}}$ & 61.99  &79.86 &54.41 &71.10\\
      &  w/ $L_{\mathrm{pgu}}$ &\textbf{66.83} & \textbf{84.16} & \textbf{59.08} & \textbf{75.92}\\
         %t2i: @R1: 0.6683, @R5: 0.8416, @R10: 0.8942, map: 0.5816
         %t2i: @R1: 0.5908, @R5: 0.7592, @R10: 0.8208, map: 0.3399      
    \hline
\end{tabu}}
\vspace{0.2cm}
\caption{
Effectiveness of the granularity-unified loss $L_{\mathrm{pgu}}$ in the T2I-ReID supervised learning.
``Single'' means that the visual and textual models are pre-trained independently.
DeiT-small \cite{touvron2021training} is the backbone.
% Ablation study on loss weight $\gamma$ in Eq.~\eqref{eq:ft_loss} in T2I-ReID supervised learning.
% When $\beta=0$, we only use the identification and ranking loss.
% When $\gamma=1$, we add the granularity-unified loss~\cite{sharma2018conceptual} for optimization.
% We set $\gamma$ to 1 by default.
% The upper/lower block shows the T2I-ReID results without/with our pre-training.
% All experiments use DeiT-small \cite{touvron2021training} as the visual backbone.
}
\vspace{-0.3cm}
\label{tab:PGU}
\end{table}

\subsection{Effectiveness of UniPT on LUPerson-T}
\myparagraph{LUPerson-T boosts the T2I-ReID performance.}
% Table~\ref{tab:luperson_t}
We compare our LUPerson-T with other generic datasets~\cite{russakovsky2015imagenet,devlin2018bert} in Table~\ref{tab:luperson_t}.
Our model, pre-trained on LUPerson-T, achieves +1.53\% and +1.48\% improvement in terms of Rank-1 accuracy on CUHK-PEDES and ICFG-PEDES with a DeiT-small backbone.
Furthermore, our pre-trained model also outperforms the generic pre-trained model on ViT-B/16 by +1.91\% and +1.47\%. 
% The underlying reason is that the visual and textual models are pre-trained separately in the previous general pre-training.
The underlying reason is that the previous generic pre-training process, under which visual and textual models are trained separately, can lead to the misalignment of the visual and textual feature representations, which harms the performance of the downstream T2I-ReID tasks. Our UniPT, based on LUPerson-T and containing pairs of images and pseudo-texts, helps to alleviate this issue to some extent.

% We show the performance improvement by replacing the pre-trained models on ImageNet\xinyu{~\cite{russakovsky2015imagenet}} and generic texts\xinyu{~\cite{hochreiter1997long, devlin2018bert}} with our visual-and-language pre-trained model on LUPerson-T on CUHK-PEDES and ICFG-PEDES.
% As shown in Table~\ref{tab:luperson_t}, based on DeiT-small, our pre-trained model have 1.53\% and 1.48\% improvement in terms of Rank-1 accuracy on two datasets.
% It also can be seen that, our pre-trained model has better performance compared with generic pre-trained model on ViT-B/16.
% Previous generic pre-training separately for the viusal and textual models may cause the misalignment of representations, bring the confusion for the optimization of downstream T2I-ReID tasks.
% Our UniPT based on LUPerson-T which contains pairs of images and pseudo texts alleviates the above problem to the some extent.

Figure \ref{fig:visualization} presents the visualization of some examples in LUPerson-T. The pseudo-texts that we generate contain basic information about each person’s characteristics. The pre-defined templates and the use of synonym replacement guarantee the fluidity and diversity of the generated pseudo-texts.

% It demonstrates that previous generic pre-training separately 

\myparagraph{UniPT improves the visual and textual representations.}
% Table~\ref{tab:pretrain}
Table~\ref{tab:pretrain} shows the effectiveness of our unified vision-and-language pre-training pipeline (UniPT).
% All experiments use the small-size ViT as as visual backbone, and the BERT as textual backbone.
In the first row in Table~\ref{tab:pretrain}, we equip the image-based pre-trained model~\cite{luo2021self} on LUPerson~\cite{fu2021unsupervised} as the visual encoder.
Compared with the top two rows, our visual pre-trained model from UniPT is largely superior than that from \cite{luo2021self} by +4.87\% and +10.90\% on CUHK-PEDES and ICFG-PEDES.
These results show that our UniPT can boost the visual representation.
Meanwhile, our UniPT can also improve the textual feature representation.
From the bottom two rows in Table~\ref{tab:pretrain}, we can see that the textual encoder from UniPT achieves +1.82\% and +0.93\% improvement compared with that from BERT~\cite{devlin2018bert} on CUHK-PEDES and ICFG-PEDES.
Therefore, our UniPT pipeline has great merit for both images and texts.

% small-size ViT which have unsupervised pre-trained~\cite{luo2021self} on LUPerson, and the Rank-1 accuracy performance drop sharply compared with the model which conduct the vision-and-language pre-training on our LUPerson-T.
% Even though replacing the visual stream with our pre-trained transformer, the performance is still unsatisfactory.
% The result demonstrate that visual-and-language pre-training is benefit to narrow the distance between visual and textual modalities.
% % Compared with the experiment of the third row which 

\begin{figure}[t!]
\begin{center}
\includegraphics[width=0.98\linewidth]{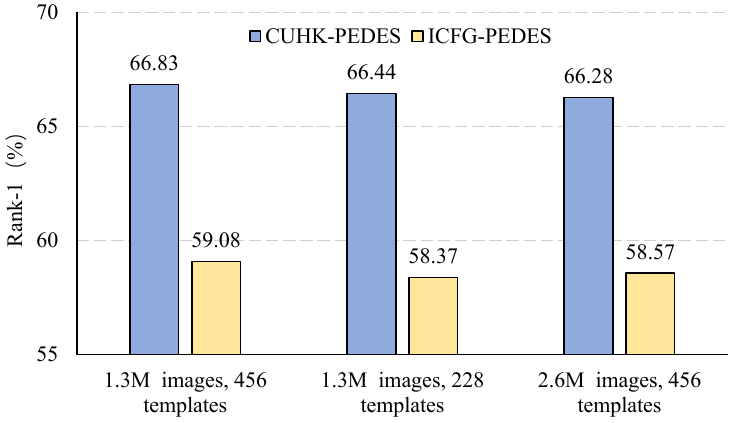}
\end{center}
\vspace{-0.3cm}
\caption{Comparison on different data scales of LUPerson-T.}
\label{fig:scale}
\vspace{-0.3cm}
\end{figure}

\myparagraph{Our UniPT \vs CLIP~\cite{radford2021learning}.}
% The most similar work with our UniPT is CLIP~\cite{radford2021learning}.
CLIP~\cite{radford2021learning} is also a vision-and-language pre-training framework, which is trained on a significant number of generic image-text pairs (400M private data pairs).
As shown in Table~\ref{tab:ablation2}, directly using the pre-trained CLIP visual and textual models as the initial visual and textual encoders can achieve comparable results.
It is because that the generic image-text pre-training is able to gain benefit from the training consistency with T2I-ReID, \ie, both tasks align the visual and textual features.
However, the data in CLIP still suffer due to the existence of a domain gap between the pre-training and the T2I-ReID task.
Compared on the ViT-B/16 backbone, our UniPT achieves 68.50\% Rank-1 accuracy while CLIP achieves only 66.34\% on CUHK-PEDES.
It is noted that even though the scale of data in UniPT is significantly smaller than that of the data in CLIP, the image–text pairs in the former are specific for persons. Therefore, the pre-trained models in our UniPT are better than those in CLIP.

% CLIP sees a large number of image-text pairs during training, which ensures its strong ability to handle cross-modal tasks.
% However, facing downstream task which contains high-quality complex descriptions for pedestrian, it is difficult to ignore the gap between pre-training datasets and T2I-ReID datasets.
% As illustrated in Table~\ref{tab:ablation2}, our UniPT with ViT-B/16 outperforms CLIP (ViT-B/16) by +2.16\% and +1.08\% on CUHK-PEDES and ICFG-PEDES, respectively.
% Considering that CLIP has no version based on DeiT-small, we adopt the version of ResNet50 which has a similar number of parameters to DeiT-small.
% It can see that, our UniPT exceed the CLIP-ResNet50 a lot which also demonstrates the effectiveness of our visual-and-language per-training.

\subsection{Ablation Study}\label{sec:ablation}
% \myparagraph{Loss weight $\beta$ in Eq.~\eqref{eq:overall_loss}.}
\myparagraph{Effectiveness of MLM loss in the pre-training.}
% Different from CLIP, our method 
Our UniPT adopts not only the contrastive loss but also the mask language modeling (MLM) objective during the pre-training process. 
We evaluate the effectiveness of MLM loss in Table~\ref{tab:mlm}.
When MLM is removed (\ie, $\beta=0$ in Eq.~\eqref{eq:ft_loss}),
the performance drops by 1.06\% and 1.73\% compared with using MLM on CUHK-PEDES and ICFG-PEDES, respectively.
The reason may be that the text descriptions are not sufficiently diverse.
The MLM makes the training harder by randomly masking some text tokens, which helps to prevent over-fitting.

% MLM loss can make the model learning harder to prevent the underlying overfitting.
% It demonstrate that MLM can make the model learn better textual representations and therefore promotes the retrieval accuracy.

% Compared with the result of subtract MLM ($\beta=0$), the final Rank-1 accuracy has improved by +1.06\% and +1.73\% on CUHK-PEDES and ICFG-PEDES, as illustrated in Table~\ref{tab:mlm}.
% This result demonstrated that MLM well guide the model to learn the higher quality textual representations and therefore promotes the retrieval accuracy.

\begin{table*}[t!]
\centering
\small
\setlength{\tabcolsep}{0.95mm}
\begin{tabular} {l|c|ccc|ccc|ccc}
    \hline
      \multirow{2}*{Methods}  & \multirow{2}*{Backbone} & \multicolumn{3}{c|}{CUHK-PEDES} & \multicolumn{3}{c|}{ICFG-PEDES}& \multicolumn{3}{c}{RSTPReid} \\
    \cline{3-11}
     & & Rank-1 & Rank-5 & Rank-10 & Rank-1 & Rank-5 & Rank-10 & Rank-1 & Rank-5 & Rank-10\\
     \hline
     \hline
      % \multirow{9}*{\rotatebox{90}{global}}
       Dual Path \cite{zheng2020dual} &ResNet50 & 44.40 & 66.26 & 75.07 & 38.99 & 59.44 & 68.41&-&-&-\\
       CMPM/C \cite{zhang2018deep} &ResNet50 & 49.37 & - & 79.27 & 43.51 & 65.44 & 74.26&-&-&-\\
       A-GANet \cite{liu2019deep} &ResNet50 & 53.14 & 74.03 & 82.95 &-&-&-&-&-&-\\
       TDE \cite{niu2020textual} &ResNet50 & 55.25 & 77.46 & 84.56&-&-&-&-&-&-\\
       VTA \cite{ge2019visual} &ResNet50 & 55.32 & 77.00 & 84.26&-&-&-&-&-&-\\
      % & CMP\_adv+TC\&IC \cite{wu2021lapscore} & 57.00 & - & 85.62\\
      % & IVT w/o MLA  \cite{shu2023see} & 62.88 & 81.60 & 87.54\\
      % \cline{2-5}
      % & \bfseries {Ours}  & {\bfseries 65.55} & {\bfseries 82.76} & {\bfseries 88.69}\\
    % \hline
      % \multirow{15}*{\rotatebox{90}{local}}
    %   & CMPM/C \cite{zhang2018deep} & 15.84 & 29.92 & 39.13\\
        MIA \cite{niu2020improving} &ResNet50 &53.10 & 75.00 & 82.90 &46.49 & 67.14 & 75.18&-&-&-\\
        SCAN \cite{lee2018stacked} &ResNet50 & 55.86 & 75.97 & 83.69 & 50.05 & 69.65 & 77.21&-&-&-\\
        ViTAA \cite{wang2020vitaa} &ResNet50 & 55.97 & 75.84 & 83.52 & 50.98 & 68.79 & 75.78&-&-&-\\
        % CMAAM \cite{aggarwal2020text} &- & 56.68 & 77.18 & 84.86\\
         HGAN \cite{zheng2020hierarchical} &ResNet50 & 59.00 & 79.49 & 86.62&-&-&-&-&-&-\\
         NAFS \cite{gao2021contextual}&ResNet50 & 59.94 & 79.86 & 86.70&-&-&-&-&-&-\\
         DSSL \cite{zhu2021dssl} &ResNet50 & 59.98 & 80.41 & 87.56 &-&-&- &32.43 &55.08 &63.19\\
         MGEL \cite{wang2021text} &ResNet50 &  60.27 & 80.01 & 86.74&-&-&-&-&-&-\\
         SSAN \cite{ding2021semantically} &ResNet50 & 61.37 & 80.15 & 86.73 & 54.23 & 72.63 & 79.53 & 43.50 & 67.80 & 77.15\\
         LapsCore \cite{wu2021lapscore} &ResNet50 & 63.40 & - & 87.80&-&-&-&-&-&-\\
         LBUL \cite{wang2022look}&ResNet50 & 64.04 & 82.66 & 87.22&-&-&- & 45.55 & 68.20 & 77.85\\
         CAIBC \cite{wang2022caibc} &ResNet50 & 64.43 & 82.87 & 88.37&-&-&- & 47.35 & 69.55 & 79.00\\
         LGUR \cite{shao2022learning} &ResNet50 &64.21 & 81.94 & 87.93 &57.42 &74.97 &81.45 &46.95 &69.90 &79.12\\
         AXM-Net \cite{farooq2022axm}&- & 64.44 & 80.52 & 86.77&-&-&-&-&-&-\\
         LGUR \cite{shao2022learning} &DeiT-small &65.25 & 83.12 & 89.00  &59.02 &75.32 &81.56 &47.95 &71.85 &80.25\\
         IVT \cite{shu2023see}&ViT-B/16 & 65.59 & 83.11 & 89.21 &56.04 &73.60 &80.22 &46.70 &70.00 &78.80\\
         \hline
         {\bfseries Our UniPT} &DeiT-small & {\bfseries66.83} & {\bfseries84.16} & {\bfseries89.42} & {\bfseries 59.08} & {\bfseries 75.92} & {\bfseries 82.08} & {\bfseries 49.45} & {\bfseries 72.75} & {\bfseries 80.35}\\
         {\bfseries Our UniPT}  &ViT-B/16 & {\bfseries68.50} & {\bfseries84.67} & {\bfseries90.38} & {\bfseries 60.09} & {\bfseries 76.19} & {\bfseries 82.46} & {\bfseries 51.85} & {\bfseries 74.85} & {\bfseries 82.85}\\
 
   %t2i: @R1: 0.6683, @R5: 0.8416, @R10: 0.8942, map: 0.5816
    % \hline 
    \hline
\end{tabular}
\vspace{0.2cm}
\caption{Performance comparisons on supervised T2I-ReID tasks on CUHK-PEDES~\cite{li2017person}, ICFG-PEDES~\cite{ding2021semantically} and RSTPReid~\cite{zhu2021dssl}.}
\vspace{-0.3cm}
\label{tab:cuhkCompare}
\end{table*}

% \myparagraph{Loss weight $\gamma$ in Eq.~\eqref{eq:ft_loss}.}
\myparagraph{Effectiveness of the granularity-unified loss in the T2I-ReID supervised learning.}
The granularity-unified loss $L_{\mathrm{pgu}}$ with the prototype-based granularity unification (PGU) module is proposed by \cite{shao2022learning} to extract granularity-unified features.
In Eq.~\eqref{eq:ft_loss}, when $\gamma=0$, the model does not use $L_{\mathrm{pgu}}$.
We evaluate the effectiveness of $L_{\mathrm{pgu}}$ on the previous separate pre-training and our UniPT.
Table~\ref{tab:PGU} shows that $L_{\mathrm{pgu}}$ can effectively improve the feature representation.

% When $\gamma=1$ in Eq.~\eqref{eq:ft_loss}, we add the granularity-unified loss for optimization.
% As shown in Table~\ref{tab:PGU}, whatever with or without our pre-training, the granularity-unified loss brings improvement, \eg, +4.84\% and +4.74\% Rank-1 acuracy on CUHK-PEDES.

\begin{table}
\centering
\small
\setlength{\tabcolsep}{0.95mm}
\begin{tabular}{l|l|ccc}
    \hline
      \multicolumn{2}{c|}{Methods} &Rank-1 & Rank-5 & Rank-10 \\
    \hline
    \hline
      \multirow{6}*{\rotatebox{90}{C $\rightarrow$ I}}
      & Dual Path \cite{zheng2020dual} & 15.41 & 29.80 & 38.19\\
    %   & CMPM/C \cite{zhang2018deep} & 15.84 & 29.92 & 39.13\\
      & MIA \cite{niu2020improving} & 19.35 & 36.78 & 46.42\\
      & SCAN \cite{lee2018stacked} & 21.27 & 39.26 & 48.83\\
      & SSAN \cite{ding2021semantically} & 24.72 & 43.43 & 53.01\\
      & SSAN(w/ BERT) \cite{ding2021semantically} & 29.24 & 49.00 & 58.53\\
      & LGUR \cite{shao2022learning} &34.25 &52.58 &60.85 \\
      \cline{2-5}
      & {\bfseries Our UniPT} (DeiT-small) &{\bfseries 35.51} &{\bfseries 54.25} &{\bfseries 62.69} \\
    % t2i: @R1: 0.3551, @R5: 0.5425, @R10: 0.6269, map: 0.164
    \hline\hline
      \multirow{6}*{\rotatebox{90}{I $\rightarrow$ C}}
        & Dual Path \cite{zheng2020dual} & 7.63 & 17.14 & 23.52\\
    %   & CMPM/C \cite{zhang2018deep} & 9.55 & 22.73 & 31.22\\
      & MIA \cite{niu2020improving} & 10.93 & 23.77 & 32.39\\
      & SCAN \cite{lee2018stacked} & 13.63 & 28.61 & 37.05\\
      & SSAN \cite{ding2021semantically} & 16.68 & 33.84 & 43.00\\
      & SSAN(w/ BERT) \cite{ding2021semantically} & 21.07 & 38.94 & 48.54\\
      & LGUR \cite{shao2022learning} &25.44 &44.48 &54.39 \\
      \cline{2-5}
      & {\bfseries Our UniPT} (DeiT-small) &{\bfseries 28.48} &{\bfseries 48.02} &{\bfseries 57.65}\\
    \hline
\end{tabular}
\vspace{0.2cm}
\caption{Performance comparison on the cross-domain setting.}
\vspace{-0.5cm}
\label{tab:crossdomain}
\end{table}

\myparagraph{Data scale.}
We provide the performance on various scales of data in Figure~\ref{fig:scale}.
As can be seen from the left two bars, a higher number of templates improves the performance due to the increase in the diversity of text descriptions.
In contrast, including more images with the same number of templates (right two bars) produces the opposite effect. There are two key reasons for this. The first is that the templates and attributes are not sufficiently abundant, meaning that many images might share the same description. The second is that selected attribute phrases are not exactly accurate; thus, more images will result in more noisy text descriptions. We thus use 1.3M image–text pairs with 456 templates by default.

\subsection{Comparison with State-of-the-Art Methods}
\myparagraph{Comparison on supervised T2I-ReID tasks.}
As illustrated in Table~\ref{tab:cuhkCompare}, our UniPT outperforms the state-of-the-art methods on CUHK-PEDES~\cite{li2017person}, ICFG-PEDES~\cite{ding2021semantically} and RSTPReid~\cite{zhu2021dssl}.
We first compare the performance of our method on CUHK-PEDES, which achieves 66.83\% and 68.50\% Rank-1 accuracy based on DeiT-small and ViT-B/16. 
LGUR~\cite{shao2022learning} proposed two modules, \ie, a dictionary-based granularity alignment module and a
prototype-based granularity unification (PGU) module, to learn granularity-unified representations for both modalities.
Here, we only equip the PGU module on the supervised learning on downstream tasks to improve the feature representation.
Our method outperforms LGUR \cite{shao2022learning} by +1.58\% in terms of Rank-1 accuracy.
IVT \cite{shu2023see} proposed multi-level alignmen (MLA) and bidirectional mask modeling (BMM), which enable the model to mine finer and more semantic alignments. 
Our method outperforms IVT by +2.91\% in terms of Rank-1 accuracy.

We further evaluate our proposed UniPT on ICFG-PEDES~\cite{ding2021semantically} and RSTPReid~\cite{zhu2021dssl}.
On ICFG-PEDES, our method achieves 59.08\% and 60.09\% Rank-1 accuracy respectively based on DeiT-small and ViT-B/16.
On RSTPReid, in a fair comparison, our method outperforms LGUR~\cite{shao2022learning} by +1.5\% with a DeiT-small backbone and surpasses IVT~\cite{shu2023see} by +5.15\% with a ViT-B/16 backbone.
In addition, the performance of our method is superior to that of the previous works LBUL \cite{wang2022look} and CAIBC~\cite{wang2022caibc}, by a large margin.

\myparagraph{Comparison on the domain generalization.}
Our model is pre-trained on the proposed LUPerson-T, which contains a large amount of knowledge derived from various images of and texts describing pedestrians.
Based on this, we conjecture that the model might have a good generalization ability.
% the domain generalization of the model will excellent.
To verify this, we conduct domain generalization experiments on two cross-domain settings, following \cite{shao2022learning}: CUHK-PEDES to ICFG-PEDES (C$\rightarrow$I) and ICFG-PEDES to CUHK-PEDES (I$\rightarrow$C).
% Specifically, we follows two cross-domain settings in \cite{shao2022learning}. 
As shown in Table~\ref{tab:crossdomain}, our architecture, pre-trained on LUPerson-T, achieves a substantial improvement compared with other methods.
For example, our model outperforms SSAN (w/ BERT)~\cite{ding2021semantically} by +6.27\%, and +7.41\% Rank-1 accuracy on C$\rightarrow$I and I$\rightarrow$C, respectively. 
Moreover, compared with LGUR~\cite{shao2022learning} which is dedicated to narrowing the granularity gap between the textual and visual feature, we achieve better accuracy overall: +1.26\% and +3.04\% on C$\rightarrow$I and I$\rightarrow$C.
% It is worth moting that, the images in the CUHK-PEDES are more diverse in style than those in the ICFG-PEDES (ICFG-PEDES are from the MSMT17 but CUHK-PDES are from the other five existing person re-identification datasets).
These experiments accordingly demonstrate the domain generalization ability of our methods.

\section{Conclusion}

This paper first reveals the data and training inconsistencies between the previous pre-training task and the T2I-ReID task.
To solve this problem, we propose a unified pre-training paradigm named UniPT for the T2I-ReID task.
Specifically, we build a text-labeled person dataset named LUPerson-T, in which pseudo-textual descriptions are automatically generated.
Moreover, we apply the vision-and-language pre-training paradigm for the model pre-training.
Our UniPT shares similar format with the T2I-ReID task from both data and training scheme perspectives.
Extensive experiments on three benchmarks verify the effectiveness of our proposed UniPT.

\vspace{0.5cm}
\noindent\textbf{Acknowledgement.}
This work was supported in part by the National Natural Science Foundation of China under Grants  62076101 and 61702193, in part by CCF-Baidu Open Fund, in part by Guangdong Basic and Applied Basic Research Foundation under Grant 2023A1515010007, in part by the Guangdong Provincial Key Laboratory of Human Digital Twin under Grant 2022B1212010004, and in part by the Program for Guangdong Introducing Innovative and Entrepreneurial Teams under Grant 2017ZT07X183.

% vision-language pre-training paradigm for text-to-image ReID task, which contains a new visual-language pre-training dataset for pedestrian, named LUPerson-T and an architecture which employs one shared Transformer encoder to extract both modalities features.
% We pretrain our architeture on LUPerson-T, and conduct
% extensive experiments on three databases to demonstrate the effectiveness of our architecture and pre-training.

% \clearpage
{\small
\bibliographystyle{ieee_fullname}
\bibliography{egbib}

\begin{thebibliography}{10}\itemsep=-1pt

\bibitem{openaiclip}
https://github.com/openai/clip.

\bibitem{chen2018improving}
Tianlang Chen, Chenliang Xu, and Jiebo Luo.
\newblock Improving text-based person search by spatial matching and adaptive
  threshold.
\newblock In {\em WACV}, pages 1879--1887. IEEE, 2018.

\bibitem{chen2020uniter}
Yen-Chun Chen, Linjie Li, Licheng Yu, Ahmed El~Kholy, Faisal Ahmed, Zhe Gan, Yu
  Cheng, and Jingjing Liu.
\newblock Uniter: Universal image-text representation learning.
\newblock In {\em ECCV}, pages 104--120. Springer, 2020.

\bibitem{devlin2018bert}
Jacob Devlin, Ming-Wei Chang, Kenton Lee, and Kristina Toutanova.
\newblock Bert: Pre-training of deep bidirectional transformers for language
  understanding.
\newblock {\em arXiv preprint arXiv:1810.04805}, 2018.

\bibitem{ding2021semantically}
Zefeng Ding, Changxing Ding, Zhiyin Shao, and Dacheng Tao.
\newblock Semantically self-aligned network for text-to-image part-aware person
  re-identification.
\newblock {\em arXiv preprint arXiv:2107.12666}, 2021.

\bibitem{dosovitskiy2020image}
Alexey Dosovitskiy, Lucas Beyer, Alexander Kolesnikov, Dirk Weissenborn,
  Xiaohua Zhai, Thomas Unterthiner, Mostafa Dehghani, Matthias Minderer, Georg
  Heigold, Sylvain Gelly, et~al.
\newblock An image is worth 16x16 words: Transformers for image recognition at
  scale.
\newblock {\em arXiv preprint arXiv:2010.11929}, 2020.

\bibitem{faghri2017vse++}
Fartash Faghri, David~J Fleet, Jamie~Ryan Kiros, and Sanja Fidler.
\newblock Vse++: Improving visual-semantic embeddings with hard negatives.
\newblock {\em arXiv preprint arXiv:1707.05612}, 2017.

\bibitem{farooq2022axm}
Ammarah Farooq, Muhammad Awais, Josef Kittler, and Syed~Safwan Khalid.
\newblock Axm-net: Implicit cross-modal feature alignment for person
  re-identification.
\newblock In {\em AAAI}, volume~36, pages 4477--4485, 2022.

\bibitem{fu2021unsupervised}
Dengpan Fu, Dongdong Chen, Jianmin Bao, Hao Yang, Lu Yuan, Lei Zhang, Houqiang
  Li, and Dong Chen.
\newblock Unsupervised pre-training for person re-identification.
\newblock In {\em CVPR}, pages 14750--14759, 2021.

\bibitem{fu2022large}
Dengpan Fu, Dongdong Chen, Hao Yang, Jianmin Bao, Lu Yuan, Lei Zhang, Houqiang
  Li, Fang Wen, and Dong Chen.
\newblock Large-scale pre-training for person re-identification with noisy
  labels.
\newblock In {\em CVPR}, pages 2476--2486, 2022.

\bibitem{fu2019horizontal}
Yang Fu, Yunchao Wei, Yuqian Zhou, Honghui Shi, Gao Huang, Xinchao Wang,
  Zhiqiang Yao, and Thomas Huang.
\newblock Horizontal pyramid matching for person re-identification.
\newblock In {\em AAAI}, volume~33, pages 8295--8302, 2019.

\bibitem{gao2021contextual}
Chenyang Gao, Guanyu Cai, Xinyang Jiang, Feng Zheng, Jun Zhang, Yifei Gong, Pai
  Peng, Xiaowei Guo, and Xing Sun.
\newblock Contextual non-local alignment over full-scale representation for
  text-based person search.
\newblock {\em arXiv preprint arXiv:2101.03036}, 2021.

\bibitem{ge2019visual}
Jing Ge, Guangyu Gao, and Zhen Liu.
\newblock Visual-textual association with hardest and semi-hard negative pairs
  mining for person search.
\newblock {\em arXiv preprint arXiv:1912.03083}, 2019.

\bibitem{he2016deep}
Kaiming He, Xiangyu Zhang, Shaoqing Ren, and Jian Sun.
\newblock Deep residual learning for image recognition.
\newblock In {\em CVPR}, pages 770--778, 2016.

\bibitem{he2021transreid}
Shuting He, Hao Luo, Pichao Wang, Fan Wang, Hao Li, and Wei Jiang.
\newblock Transreid: Transformer-based object re-identification.
\newblock In {\em ICCV}, pages 15013--15022, 2021.

\bibitem{hochreiter1997long}
Sepp Hochreiter and J{\"u}rgen Schmidhuber.
\newblock Long short-term memory.
\newblock {\em Neural computation}, 9(8):1735--1780, 1997.

\bibitem{jing2020pose}
Ya Jing, Chenyang Si, Junbo Wang, Wei Wang, Liang Wang, and Tieniu Tan.
\newblock Pose-guided multi-granularity attention network for text-based person
  search.
\newblock In {\em AAAI}, 2020.

\bibitem{kingma2014adam}
Diederik~P Kingma and Jimmy Ba.
\newblock Adam: A method for stochastic optimization.
\newblock {\em ICLR}, 2015.

\bibitem{lee2018stacked}
Kuang-Huei Lee, Xi Chen, Gang Hua, Houdong Hu, and Xiaodong He.
\newblock Stacked cross attention for image-text matching.
\newblock In {\em ECCV}, pages 201--216, 2018.

\bibitem{li2020unicoder}
Gen Li, Nan Duan, Yuejian Fang, Ming Gong, and Daxin Jiang.
\newblock Unicoder-vl: A universal encoder for vision and language by
  cross-modal pre-training.
\newblock In {\em AAAI}, volume~34, pages 11336--11344, 2020.

\bibitem{li2019visualbert}
Liunian~Harold Li, Mark Yatskar, Da Yin, Cho-Jui Hsieh, and Kai-Wei Chang.
\newblock Visualbert: A simple and performant baseline for vision and language.
\newblock {\em arXiv preprint arXiv:1908.03557}, 2019.

\bibitem{li2017identity}
Shuang Li, Tong Xiao, Hongsheng Li, Wei Yang, and Xiaogang Wang.
\newblock Identity-aware textual-visual matching with latent co-attention.
\newblock In {\em ICCV}, pages 1890--1899, 2017.

\bibitem{li2017person}
Shuang Li, Tong Xiao, Hongsheng Li, Bolei Zhou, Dayu Yue, and Xiaogang Wang.
\newblock Person search with natural language description.
\newblock In {\em CVPR}, pages 1970--1979, 2017.

\bibitem{liu2020graph}
Chunxiao Liu, Zhendong Mao, Tianzhu Zhang, Hongtao Xie, Bin Wang, and Yongdong
  Zhang.
\newblock Graph structured network for image-text matching.
\newblock In {\em CVPR}, pages 10921--10930, 2020.

\bibitem{liu2019deep}
Jiawei Liu, Zheng-Jun Zha, Richang Hong, Meng Wang, and Yongdong Zhang.
\newblock Deep adversarial graph attention convolution network for text-based
  person search.
\newblock In {\em ACMMM}, pages 665--673, 2019.

\bibitem{loshchilov2017decoupled}
Ilya Loshchilov and Frank Hutter.
\newblock Decoupled weight decay regularization.
\newblock {\em arXiv preprint arXiv:1711.05101}, 2017.

\bibitem{lu2019vilbert}
Jiasen Lu, Dhruv Batra, Devi Parikh, and Stefan Lee.
\newblock Vilbert: Pretraining task-agnostic visiolinguistic representations
  for vision-and-language tasks.
\newblock {\em arXiv preprint arXiv:1908.02265}, 2019.

\bibitem{luo2021self}
Hao Luo, Pichao Wang, Yi Xu, Feng Ding, Yanxin Zhou, Fan Wang, Hao Li, and Rong
  Jin.
\newblock Self-supervised pre-training for transformer-based person
  re-identification.
\newblock {\em arXiv preprint arXiv:2111.12084}, 2021.

\bibitem{niu2020improving}
Kai Niu, Yan Huang, Wanli Ouyang, and Liang Wang.
\newblock Improving description-based person re-identification by
  multi-granularity image-text alignments.
\newblock {\em TIP}, 29:5542--5556, 2020.

\bibitem{niu2020textual}
Kai Niu, Yan Huang, and Liang Wang.
\newblock Textual dependency embedding for person search by language.
\newblock In {\em ACMMM}, pages 4032--4040, 2020.

\bibitem{oord2018representation}
Aaron van~den Oord, Yazhe Li, and Oriol Vinyals.
\newblock Representation learning with contrastive predictive coding.
\newblock {\em arXiv preprint arXiv:1807.03748}, 2018.

\bibitem{radford2021learning}
Alec Radford, Jong~Wook Kim, Chris Hallacy, Aditya Ramesh, Gabriel Goh,
  Sandhini Agarwal, Girish Sastry, Amanda Askell, Pamela Mishkin, Jack Clark,
  et~al.
\newblock Learning transferable visual models from natural language
  supervision.
\newblock In {\em ICML}, pages 8748--8763. PMLR, 2021.

\bibitem{russakovsky2015imagenet}
Olga Russakovsky, Jia Deng, Hao Su, Jonathan Krause, Sanjeev Satheesh, Sean Ma,
  Zhiheng Huang, Andrej Karpathy, Aditya Khosla, and Michael Bernstein.
\newblock Imagenet large scale visual recognition challenge.
\newblock {\em IJCV}, 115(3):211--252, 2015.

\bibitem{sarafianos2019adversarial}
Nikolaos Sarafianos, Xiang Xu, and Ioannis~A Kakadiaris.
\newblock Adversarial representation learning for text-to-image matching.
\newblock In {\em ICCV}, pages 5814--5824, 2019.

\bibitem{shao2022learning}
Zhiyin Shao, Xinyu Zhang, Meng Fang, Zhifeng Lin, Jian Wang, and Changxing
  Ding.
\newblock Learning granularity-unified representations for text-to-image person
  re-identification.
\newblock In {\em ACMMM}, pages 5566--5574, 2022.

\bibitem{shu2023see}
Xiujun Shu, Wei Wen, Haoqian Wu, Keyu Chen, Yiran Song, Ruizhi Qiao, Bo Ren,
  and Xiao Wang.
\newblock See finer, see more: Implicit modality alignment for text-based
  person retrieval.
\newblock In {\em ECCVW}, pages 624--641. Springer, 2023.

\bibitem{su2019vl}
Weijie Su, Xizhou Zhu, Yue Cao, Bin Li, Lewei Lu, Furu Wei, and Jifeng Dai.
\newblock Vl-bert: Pre-training of generic visual-linguistic representations.
\newblock {\em arXiv preprint arXiv:1908.08530}, 2019.

\bibitem{sun2019videobert}
Chen Sun, Austin Myers, Carl Vondrick, Kevin Murphy, and Cordelia Schmid.
\newblock Videobert: A joint model for video and language representation
  learning.
\newblock In {\em ICCV}, pages 7464--7473, 2019.

\bibitem{sun2018beyond}
Yifan Sun, Liang Zheng, Yi Yang, Qi Tian, and Shengjin Wang.
\newblock Beyond part models: Person retrieval with refined part pooling (and a
  strong convolutional baseline).
\newblock In {\em ECCV}, pages 480--496, 2018.

\bibitem{tan2019lxmert}
Hao Tan and Mohit Bansal.
\newblock Lxmert: Learning cross-modality encoder representations from
  transformers.
\newblock {\em arXiv preprint arXiv:1908.07490}, 2019.

\bibitem{touvron2021training}
Hugo Touvron, Matthieu Cord, Matthijs Douze, Francisco Massa, Alexandre
  Sablayrolles, and Herv{\'e} J{\'e}gou.
\newblock Training data-efficient image transformers \& distillation through
  attention.
\newblock In {\em ICML}, pages 10347--10357. PMLR, 2021.

\bibitem{wang2021text}
Chengji Wang, Zhiming Luo, Yaojin Lin, and Shaozi Li.
\newblock Text-based person search via multi-granularity embedding learning.
\newblock In {\em IJCAI}, pages 1068--1074, 2021.

\bibitem{wang2022image}
Wenhui Wang, Hangbo Bao, Li Dong, Johan Bjorck, Zhiliang Peng, Qiang Liu, Kriti
  Aggarwal, Owais~Khan Mohammed, Saksham Singhal, Subhojit Som, et~al.
\newblock Image as a foreign language: Beit pretraining for all vision and
  vision-language tasks.
\newblock {\em arXiv preprint arXiv:2208.10442}, 2022.

\bibitem{wang2019language}
Yuyu Wang, Chunjuan Bo, Dong Wang, Shuang Wang, Yunwei Qi, and Huchuan Lu.
\newblock Language person search with mutually connected classification loss.
\newblock In {\em ICASSP}, pages 2057--2061. IEEE, 2019.

\bibitem{wang2020vitaa}
Zhe Wang, Zhiyuan Fang, Jun Wang, and Yezhou Yang.
\newblock Vitaa: Visual-textual attributes alignment in person search by
  natural language.
\newblock In {\em ECCV}, pages 402--420. Springer, 2020.

\bibitem{wang2022caibc}
Zijie Wang, Aichun Zhu, Jingyi Xue, Xili Wan, Chao Liu, Tian Wang, and Yifeng
  Li.
\newblock Caibc: Capturing all-round information beyond color for text-based
  person retrieval.
\newblock In {\em ACMMM}, pages 5314--5322, 2022.

\bibitem{wang2022look}
Zijie Wang, Aichun Zhu, Jingyi Xue, Xili Wan, Chao Liu, Tian Wang, and Yifeng
  Li.
\newblock Look before you leap: Improving text-based person retrieval by
  learning a consistent cross-modal common manifold.
\newblock In {\em ACMMM}, pages 1984--1992, 2022.

\bibitem{wu2021lapscore}
Yushuang Wu, Zizheng Yan, Xiaoguang Han, Guanbin Li, Changqing Zou, and
  Shuguang Cui.
\newblock Lapscore: Language-guided person search via color reasoning.
\newblock In {\em ICCV}, pages 1624--1633, 2021.

\bibitem{yang2022unleashing}
Zizheng Yang, Xin Jin, Kecheng Zheng, and Feng Zhao.
\newblock Unleashing potential of unsupervised pre-training with intra-identity
  regularization for person re-identification.
\newblock In {\em CVPR}, pages 14298--14307, 2022.

\bibitem{you2022learning}
Haoxuan You, Luowei Zhou, Bin Xiao, Noel Codella, Yu Cheng, Ruochen Xu, Shih-Fu
  Chang, and Lu Yuan.
\newblock Learning visual representation from modality-shared contrastive
  language-image pre-training.
\newblock In {\em ECCV}, pages 69--87. Springer, 2022.

\bibitem{zhang2022negative}
Kun Zhang, Zhendong Mao, Quan Wang, and Yongdong Zhang.
\newblock Negative-aware attention framework for image-text matching.
\newblock In {\em CVPR}, pages 15661--15670, 2022.

\bibitem{zhang2019self}
Xinyu Zhang, Jiewei Cao, Chunhua Shen, and Mingyu You.
\newblock Self-training with progressive augmentation for unsupervised
  cross-domain person re-identification.
\newblock In {\em ICCV}, pages 8222--8231, 2019.

\bibitem{zhang2020memorizing}
Xinyu Zhang, Dong Gong, Jiewei Cao, and Chunhua Shen.
\newblock Memorizing comprehensively to learn adaptively: Unsupervised
  cross-domain person re-id with multi-level memory.
\newblock {\em arXiv preprint arXiv:2001.04123}, 2020.

\bibitem{zhang2022implicit}
Xinyu Zhang, Dongdong Li, Zhigang Wang, Jian Wang, Errui Ding, Javen~Qinfeng
  Shi, Zhaoxiang Zhang, and Jingdong Wang.
\newblock Implicit sample extension for unsupervised person re-identification.
\newblock In {\em CVPR}, pages 7369--7378, 2022.

\bibitem{zhang2022contrastive}
Yuhao Zhang, Hang Jiang, Yasuhide Miura, Christopher~D Manning, and Curtis~P
  Langlotz.
\newblock Contrastive learning of medical visual representations from paired
  images and text.
\newblock In {\em MLHC}, pages 2--25. PMLR, 2022.

\bibitem{zhang2018deep}
Ying Zhang and Huchuan Lu.
\newblock Deep cross-modal projection learning for image-text matching.
\newblock In {\em ECCV}, pages 686--701, 2018.

\bibitem{zheng2020hierarchical}
Kecheng Zheng, Wu Liu, Jiawei Liu, Zheng-Jun Zha, and Tao Mei.
\newblock Hierarchical gumbel attention network for text-based person search.
\newblock In {\em ACM MM}, pages 3441--3449, 2020.

\bibitem{zheng2015scalable}
Liang Zheng, Liyue Shen, Lu Tian, Shengjin Wang, Jingdong Wang, and Qi Tian.
\newblock Scalable person re-identification: A benchmark.
\newblock In {\em ICCV}, pages 1116--1124, 2015.

\bibitem{zheng2019joint}
Zhedong Zheng, Xiaodong Yang, Zhiding Yu, Liang Zheng, Yi Yang, and Jan Kautz.
\newblock Joint discriminative and generative learning for person
  re-identification.
\newblock In {\em CVPR}, 2019.

\bibitem{zheng2020dual}
Zhedong Zheng, Liang Zheng, Michael Garrett, Yi Yang, Mingliang Xu, and Yi-Dong
  Shen.
\newblock Dual-path convolutional image-text embeddings with instance loss.
\newblock {\em TOMM}, 16(2):1--23, 2020.

\bibitem{zhong2020learning}
Zhun Zhong, Liang Zheng, Zhiming Luo, Shaozi Li, and Yi Yang.
\newblock Learning to adapt invariance in memory for person re-identification.
\newblock {\em TPAMI}, 2020.

\bibitem{zhou2020unified}
Luowei Zhou, Hamid Palangi, Lei Zhang, Houdong Hu, Jason Corso, and Jianfeng
  Gao.
\newblock Unified vision-language pre-training for image captioning and vqa.
\newblock In {\em AAAI}, volume~34, pages 13041--13049, 2020.

\bibitem{zhu2021dssl}
Aichun Zhu, Zijie Wang, Yifeng Li, Xili Wan, Jing Jin, Tian Wang, Fangqiang Hu,
  and Gang Hua.
\newblock Dssl: Deep surroundings-person separation learning for text-based
  person retrieval.
\newblock In {\em ACMMM}, pages 209--217, 2021.

\bibitem{zhu2022pass}
Kuan Zhu, Haiyun Guo, Tianyi Yan, Yousong Zhu, Jinqiao Wang, and Ming Tang.
\newblock Pass: Part-aware self-supervised pre-training for person
  re-identification.
\newblock In {\em ECCV}, pages 198--214, 2022.

\bibitem{zhu2015aligning}
Yukun Zhu, Ryan Kiros, Rich Zemel, Ruslan Salakhutdinov, Raquel Urtasun,
  Antonio Torralba, and Sanja Fidler.
\newblock Aligning books and movies: Towards story-like visual explanations by
  watching movies and reading books.
\newblock In {\em ICCV}, pages 19--27, 2015.

\end{thebibliography}
}

\end{document}